\documentclass[pagesize=auto,openany,oneside,bibliography=totocnumbered,pdftex]{scrbook}

\usepackage{comment}
\usepackage{etoolbox}
\usepackage{fancyvrb}
\usepackage{graphicx}
\usepackage{hyperref}
\usepackage[round]{natbib}
\usepackage{xspace}
\usepackage{xcolor}

\usepackage[T2A]{fontenc}
\usepackage[utf8]{inputenc}
\usepackage[greek,russian,english]{babel}
\newcommand\nextpage[1][]{
\ifdefined\HCode {
  \HCode{<mbp:pagebreak />}}
\else
  \newpage
\fi
}

\def\surl#1{{\small{\url{#1}}}}

\usepackage{graphicx}%
\usepackage{multirow}%
\usepackage{amsmath,amssymb,amsfonts}%
\usepackage{amsthm}%
\usepackage{mathrsfs}%
\usepackage[title]{appendix}%
\usepackage{xcolor}%
\usepackage{textcomp}%
\usepackage{manyfoot}%
\usepackage{booktabs}%
\usepackage{algorithm}%
\usepackage{algorithmicx}%
\usepackage{algpseudocode}%
\usepackage{listings}%
\usepackage{epigraph}
\usepackage{booktabs} 
\usepackage{geometry} 
\usepackage{tabularx} 
\usepackage{hyperref}
\usepackage{xurl}
\usepackage{array}
\usepackage{makecell}
\usepackage{pdfpages}
\usepackage{wrapfig}
\usepackage{fancyhdr}
\usepackage{tocloft}
\usepackage{setspace}

\ifdefined\HCode{\KOMAoptions{twoside=false}} \fi

\begin{document}
\setstretch{1.15} 
%\onehalfspacing % Espacio de 1.5
% Title and author are written down in the cover_page.tex file, and author
% appears also as an argument to the ebook-convert command in the build.sh file.

\includepdf[pages=1]{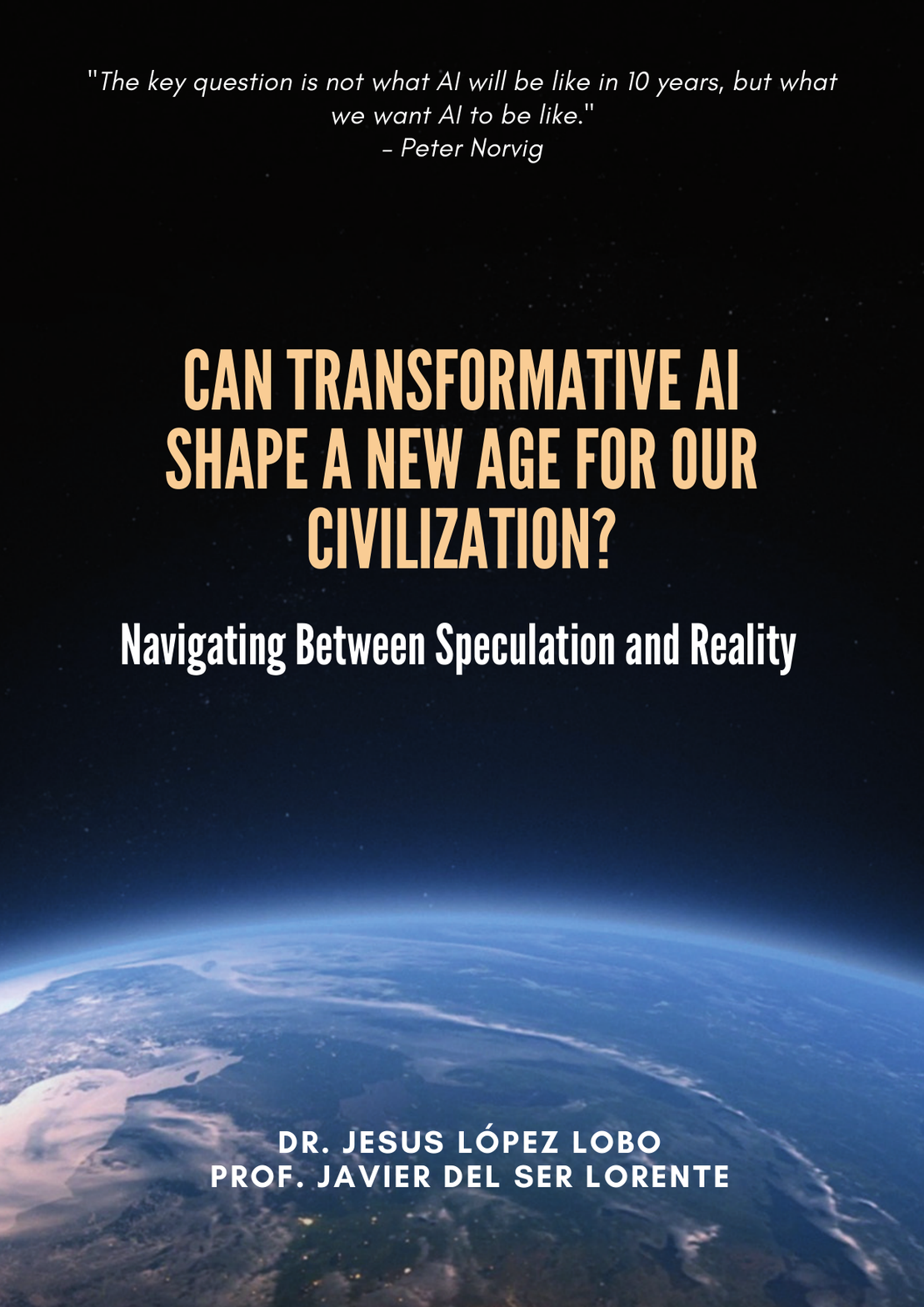}

% Añadir una página en blanco
\newpage
\thispagestyle{empty} % Evita números de página en blanco
\mbox{} % Página completamente en blanco
\nextpage
\tableofcontents
\nextpage
\listoffigures
\nextpage
\listoftables

\nextpage
%% Introduction / first chapter.

\chapter*{Abstract}

Artificial Intelligence is widely regarded as a transformative force with the potential to redefine numerous sectors of human civilization. While Artificial Intelligence has evolved from speculative fiction to a pivotal element of technological progress, its role as a truly transformative agent, or transformative Artificial Intelligence, remains a subject of debate. This work explores the historical precedents of technological breakthroughs, examining whether Artificial Intelligence can achieve a comparable impact, and it delves into various ethical frameworks that shape the perception and development of Artificial Intelligence. Additionally, it considers the societal, technical, and regulatory challenges that must be addressed for Artificial Intelligence to become a catalyst for global change. We also examine not only the strategies and methodologies that could lead to transformative Artificial Intelligence but also the barriers that could ultimately make these goals unattainable. We end with a critical inquiry into whether reaching a transformative Artificial Intelligence might compel humanity to adopt an entirely new ethical approach, tailored to the complexities of advanced Artificial Intelligence. By addressing the ethical, social, and scientific dimensions of Artificial Intelligence's development, this work contributes to the broader discourse on the long-term implications of Artificial Intelligence and its capacity to drive civilization toward a new era of progress or, conversely, exacerbate existing inequalities and risks.

\vspace{1em}
\noindent\textbf{Keywords:} Transformative AI, AI Ethics, AI Governance, AI Alignment, AI Regulation 
\nextpage
\chapter{Introduction}\label{intro}

The concept of Artificial Intelligence (AI), once a speculative idea confined to science fiction, has evolved into one of the most transformative forces shaping our modern civilization. Rooted in philosophical and mathematical inquiries into the nature of intelligence and computation, AI has grown exponentially, deeply intertwined with developments in Computer Science, Neuroscience, Maths, and Cognitive Psychology, among others. The vision of machines capable of replicating or surpassing human intelligence, first proposed by pioneers such as \citep{turing2009computing}, laid the groundwork for a broader discourse on the nature of mind and machine.

In recent decades, breakthroughs in machine learning, particularly deep learning \citep{lecun2015deep}, have shifted the landscape from theoretical speculation to tangible achievements. Systems based on generative AI \citep{banh2023generative} are demonstrating remarkable capabilities of modern AI, which now exceed human performance in specific domains such as strategy games, natural language processing, or image recognition \citep{aiindex2024}. However, the impact of AI on our civilization extends beyond its technical milestones. The integration of AI into critical social functions, from healthcare to governance, can be regarded as a double-edged sword. On the one hand, AI offers the promise of solving some of the most complex problems of humanity, such as climate modeling \citep{ripple20242024}, personalized medicine \citep{sadee2023pharmacogenomics}, or economic optimization \citep{zheng2021aieconomistoptimaleconomic}. On the other hand, it exacerbates concerns about surveillance, labor displacement, and algorithmic biases \cite{hickok2023policy,frank2023ai,ferrara2023fairness}. The need for robust ethical frameworks and governance structures has never been more urgent. Since we stand at the precipice of what might be a new era in human history, the trajectory of AI's development will shape the future of human civilization. Moreover, it is a process that has already begun, the ultimate trajectory of which remains uncertain concerning its potential impact on social development \citep{zershaaneh2022transformative}. Whether it leads to unprecedented prosperity or catastrophic outcomes will depend not only on technological advancements, but also on how societies choose to harness and regulate this formidable lever of transformative change. 

Due to these expectations that AI has recently spawn, the term Transformative AI (\texttt{TAI}) has become a topic of growing interest \citep{buccella2023ai,banafa2024transformative,rawas2024ai} due to recent breakthroughs in machine learning and neural computation, which have brought us closer to developing systems with the potential to revolutionize industries, reshape societal norms, and raise unprecedented ethical and governance challenges. The urgency surrounding \texttt{TAI} is further amplified by its potential to dramatically impact the global economy, human labor, and decision-making processes, prompting a widespread debate on its regulation and long-term societal implications. Therefore, in this work we adopt this term to refer to the development of those AI systems with the potential to induce significant and far-reaching transformations across civilization. As suggested in \citep{gruetzemacher2019defining,gruetzemacher2022transformative}, it is a good practice to embrace the term \texttt{TAI} because it encompasses the notion that a diverse range of advanced AI systems warrant attention and concern, and accounts for the possibility that certain types of advanced AI systems could have profoundly transformative effects on our society without necessarily exhibit all human or even superior capabilities. \texttt{TAI}, as opposed to other terms such as Artificial General Intelligence (AGI) \citep{goertzel2007artificial,morris2023levels}, Human-Level Intelligence (HLAI) \citep{mccarthy2007here}, Artificial Superintelligence (ASI) \citep{bostrom2014superintelligence}, or simply \textit{strong AI}, does not make broad assumptions about the characteristics that advanced AI systems must show to bring about significant societal change. Due to the existence of such diverse and non-consensual terms to refer to AI of such characteristics, we will simply use the neutral term Broadly Capable AI (\texttt{BCAI}) so as not to incur errors or inaccuracies. It is not one of the objectives of this work to make a clear distinction between them, especially when such definitions are still a subject of intense debate.

Moreover, we see \texttt{TAI} paralleling the space race in regards to its transformative potential and interdisciplinary nature. Much like the space race, which required breakthroughs across several scientific fields, AI mainly combines knowledge from several fields, including Mathematics, Statistics, Computer Science, Neuroscience, and Ethics, to tackle its own grand challenges. Both are more than technological feats; they represent a collective effort to expand the boundaries of human knowledge. As space exploration unlocked new frontiers beyond Earth, AI is poised to reshape how we interact with the world, augmenting our intellectual and creative capabilities, especially as AI consolidates its transformative potential as \texttt{TAI}.

The idea of transformative technologies is not new. History offers numerous examples of innovations that have catalyzed epochal shifts, such as the advent of the steam engine during the Industrial Revolution or the rise of the Internet in the Information Age. However, what distinguishes \texttt{TAI} from these previous breakthroughs is its potential to serve as a ``general-purpose technology'' \citep{triguero2024general} with capabilities that permeate nearly all domains of human activity. As AI research progresses, discussions increasingly focus on whether \texttt{TAI} could represent not merely a significant technological advancement, but a fundamental shift in civilization itself. While the excitement around \texttt{TAI} is palpable, so are the ethical, social, and existential challenges it poses, raising the question: will \texttt{TAI} serve as a beneficial force for humanity, or will it exacerbate existing inequalities and introduce unforeseen risks \citep{slattery2024ai}? As shown, \texttt{TAI} has been in the limelight for some time now, and the state-of-the-art has been addressing these and more questions for some time.

Some of the most relevant publications about \texttt{TAI} focus on its definition and levels \citep{gruetzemacher2019defining,gruetzemacher2022transformative}, on the challenges it addresses \citep{rawas2024ai}, and on its practical applications \citep{banafa2024transformative}. Our article provides a comprehensive overview of \texttt{TAI} within the context of our civilization. We explore its historical significance as a groundbreaking phenomenon, examine it through an ethical lens, and evaluate its feasibility from theoretical, human, and technical perspectives. The work is also updated with the latest trends, theorems, and thoughts on \texttt{TAI}. The discussion concludes with an analysis of the potential impact that \texttt{TAI} could have in years to come. Altogether, our reflections herein offered can be regarded as an extensive compendium of what might unfold if current AI-based advancements become a catalyst for a global and transformative change in our civilization.

The rest of the article proceeds first with the Chapter \ref{unders_perceiv}, where shows (1) a historical overview of major technological advancements that have reshaped human civilization, drawing parallels to the potential impact of \texttt{TAI} (Section \ref{hist}); and (2) the ethical implications of AI are examined through various philosophical lenses to analyze how different perspectives influence the development and deployment of AI systems (Section \ref{ethics}). In the Chapter \ref{part_2} we got into (1) how AI has been with us for some time now, and what it means today, showing the significance of it in our civilization, and the possible roles it is destined to play (Section \ref{here_again}); and (2) the human (Section \ref{human_limits}) and technical (Section \ref{techsci_limits}) obstacles that exist in the transformation of AI into \texttt{TAI}. After that, we present the Chapter \ref{opps_persp}, which (1) shows the opportunities of \texttt{TAI} to become a reality (Section \ref{tai_yes}); (2) it analyzes what could be the great hope of achieving \texttt{TAI} (Section \ref{the_hope}); and (3) assuming the advent of \texttt{TAI}, it highlights the possible need for a new ethical and philosophical approach to AI (Section \ref{new_rel}). The final Chapter \ref{ref_conc} (1) presents a discussion on the potential futures of AI, exploring both the opportunities for human enhancement and the risks of over-reliance on it (Section \ref{disc}); and (2) it concludes by emphasizing the need for a balanced approach to AI development, one that integrates ethical considerations, robust governance, and a clear societal vision for the role of AI in the future (Section \ref{conc}).

\chapter{Understanding and Perceiving AI }\label{unders_perceiv}
\section{A Few Brushstrokes of History}\label{hist}

\begin{figure}
    \centering
    \includegraphics[width=0.9\linewidth]{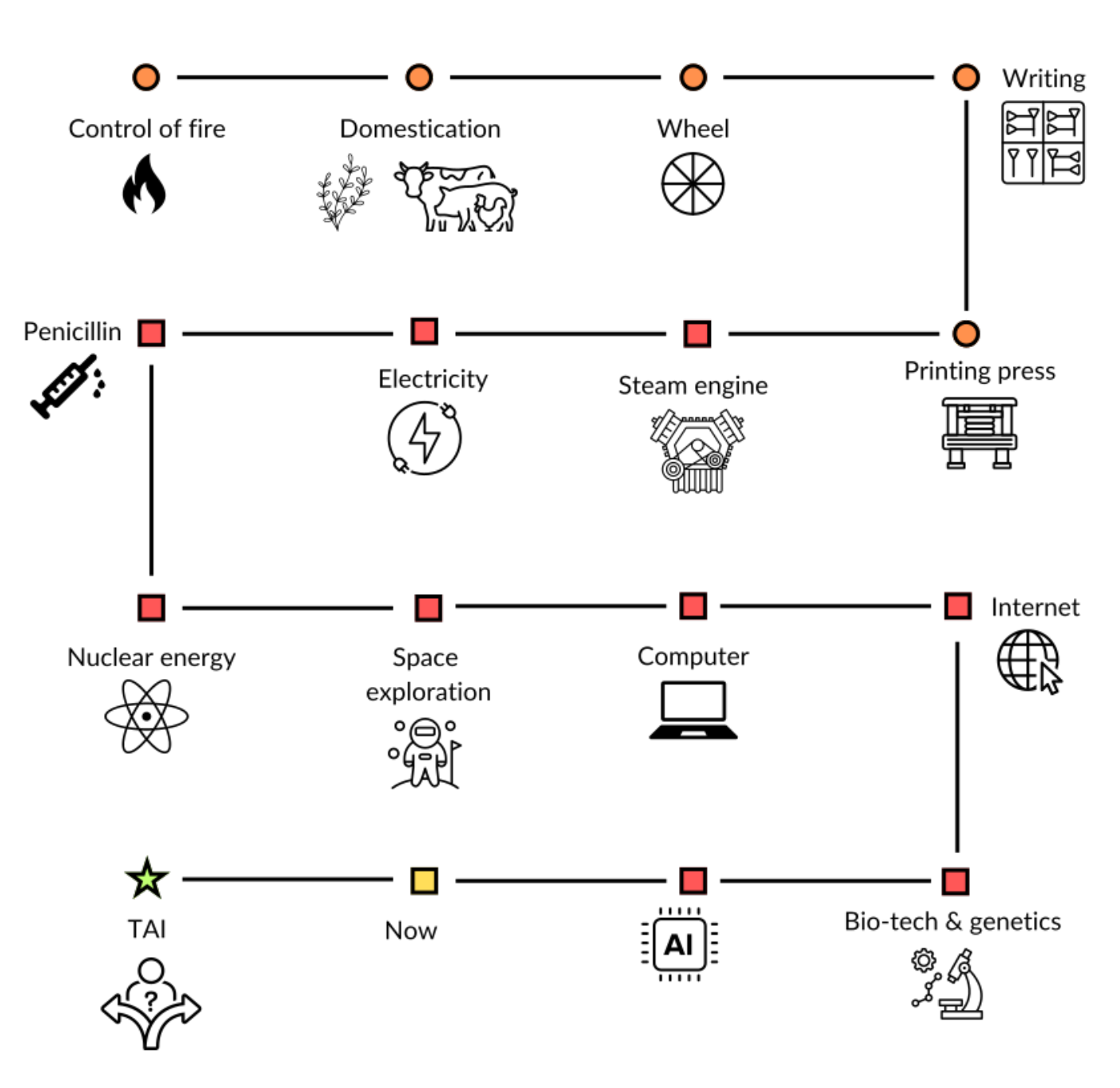}
    \caption[The harnessing of some technological advancements that pushed our civilization forward, and the uncertainty about the role of AI in this roadmap.]{The harnessing of some technological advancements that pushed our civilization forward and the uncertainty about the role of AI in this roadmap. Orange circles are used for those advancements before the Modern Age, Red squares are used for those in the Modern Age, the yellow square indicates the current moment, and the yellow star is for the possible emergence of \texttt{TAI} in the future. Note that the chronological order of such advancements is debatable from several points of view, but it has been arranged at the discretion of the authors according to the exploitation of the turning points and not their discovery.}
    \label{fig:advances}
\end{figure}

Throughout human history, certain technological advancements have marked profound turning points, pushing our civilization forward and reshaping the very fabric of society. These innovations are not just incremental improvements, but monumental leaps that have transformed how humans live, work, and interact with the world. The purpose of this section is to highlight some milestones in human history that are widely accepted, and that have represented a real leap forward for our civilization (depicted in Figure \ref{fig:advances}). We start this tour with the aim of understanding what can be expected from modern AI, and whether it can genuinely be included in this compendium of transformative milestones.

\subsection{The Early Turning Points}\label{old_turnings}

One of the earliest and most fundamental advancements was \emph{the control of fire} \citep{James1989HominidUO}, which provided warmth, protection from predators, and the ability to cook food. This discovery, in turn, had significant impacts on human nutrition and social structures. The mastery of fire was a critical step in human evolution, laying the groundwork for future developments. Another transformative achievement was \emph{the domestication of animals and plants}, which began with the dog \citep{machugh2017taming} and cereals (e.g., wheat and barley) \citep{Fuller2011CultivationAD}. This process allowed humans to transition from nomadic lifestyles to settled agricultural societies. Domestication provided not only a reliable source of food but also labor, transportation, and materials like wool and leather, enabling the growth of stable communities, the development of complex societies, specialized labor, and the development of trade networks. The introduction of \emph{the wheel} \citep{bulliet2016wheel} further revolutionized human capabilities. This seemingly simple invention greatly improved transportation and trade, facilitated the movement of goods over long distances and spurred the development of machinery, fundamentally altering the course of economic and social development. 

Differently, the invention of \emph{writing} \citep{robinson2018origins} marked the dawn of recorded history. Writing enabled the preservation and transmission of knowledge across generations, allowing for the codification of laws, the administration of governments, and the flourishing of literature, science, and philosophy, being a crucial step in the development of complex societies and cultures. Many centuries later, the \emph{printing press} \citep{eisenstein1980printing} democratized knowledge and education. By enabling the mass production of books, the printing press accelerated the spread of ideas, fueling the Renaissance, the Reformation, and the Scientific Revolution. It was instrumental in breaking the monopoly of knowledge held by the elite, allowing for broader access to information and literacy.

\subsection{The Modern Turning Points}\label{modern_turnings}

Already in the Modern Age, \emph{the steam engine} \citep{dickinson2022short} was the driving force behind the Industrial Revolution. This innovation mechanized production, leading to unprecedented increases in efficiency and productivity. It transformed industries, transportation, and society, marking the transition from agrarian economies to industrial powerhouses. Later, the discovery and harnessing of \emph{electricity} \citep{jonnes2004empires} brought about another wave of transformation. Electricity became the lifeblood of modern civilization, powering homes, industries, and communication systems. It revolutionized daily life, enabling the development of technologies that continue to shape the world today. In the realm of medicine, the discovery of \emph{penicillin} \citep{dougherty2011antibiotic} marked the beginning of the modern antibiotic era. This breakthrough has saved millions of lives by providing an effective treatment for bacterial infections, fundamentally changing medical practice and public health. 

The advent of \emph{computers} \citep{haigh2021new} ushered in the Information Age. Computers revolutionized how we process, store, and analyze information, leading to the development of the Internet, smartphones, and countless other technologies that permeate every aspect of modern life. Concretely, the \emph{Internet} \citep{leiner2009brief} is arguably one of the most transformative technologies in human history. It has connected people across the globe, enabling instant communication, the democratization of information, and the creation of new industries and forms of social interaction. The Internet has fundamentally reshaped economies, politics, and cultures worldwide. In more recent times, \emph{biotechnology} and \emph{genetic engineering} have opened new frontiers in medicine, agriculture, and beyond. The ability to manipulate DNA \citep{pickar2019next} has led to the development of life-saving therapies, genetically modified crops, and the potential for unprecedented advancements in biology and medicine. The harnessing of \emph{nuclear energy} \citep{hewitt2000introduction} provided a powerful and relatively clean source of energy, although it also introduced new challenges in terms of safety and waste management. Nonetheless, nuclear power has become a critical part of the global energy mix, offering a solution to the growing demand for electricity. The ongoing \emph{exploration of space} \citep{6174432,jiang2021avoiding} represents humanity's quest to extend its reach beyond Earth. Space exploration has not only expanded our understanding of the universe but has also led to technological innovations that have applications in everyday life, such as satellite communication and the Global Position System (GPS).

\subsection{The AI's Turn: the Great Hope}

Today, it appears to be the turn of AI, a turning point poised to revolutionize human capabilities and societal structures in ways we are only beginning to comprehend. The potential of AI to achieve \texttt{TAI} represents not just another leap in technological innovation, but a turning point that could redefine the trajectory of human progress. Unfortunately, it is impossible to anticipate precisely how this might occur. Previously, other technologies had similar disruptive potential but fell by the wayside, becoming part of a list of unfulfilled promises \cite{floridi2024ai}. Some breakthroughs have had a profound impact on the world, whereas others, despite initial promise, have failed to achieve the level of adoption or impact that was anticipated \citep{christensen2015innovator,o2021lead}. Rapid advancements in AI have already demonstrated its ability to perform tasks once thought uniquely human. However, the transition to \texttt{TAI} requires overcoming several key challenges. Next, we briefly introduce some of them to guide the reader, but they will be discussed in more depth in the chapters of this work. 

First, AI must transcend narrow domains and develop the capacity for generalization, allowing it to adapt knowledge and skills across diverse fields. This would involve significant progress in areas such as transfer learning \citep{zhu2023transfer}, unsupervised learning \citep{james2023unsupervised}, or the recent world models \citep{xiang2024language}, by which AI systems learn internal representations of their environments. Second, the development of reasoning and causal inference mechanisms is crucial. Current AI models, while powerful, primarily rely on statistical correlations rather than understanding the underlying cause-and-effect relationships within data. Embedding causal reasoning would enable AI to predict outcomes, simulate scenarios, and make decisions with greater depth and reliability, thereby closing the gap between specialized AI systems and the general-purpose reasoning required for \texttt{TAI}. Third, achieving \texttt{TAI} demands unprecedented levels of computational power and optimization. Innovations in hardware could include e.g., neuromorphic computing or quantum technologies; they would be likely to play a critical role in bridging this gap. These advancements, coupled with efficient energy utilization and scalable architectures, would provide the necessary infrastructure to support the vast computational demands of \texttt{TAI} systems. Finally, ethical and philosophical considerations must guide this pursuit. \texttt{TAI} is not merely a technological milestone but a societal one, with implications for employment, governance, security, and even the nature of humanity itself. Collaborative global efforts that incorporate interdisciplinary perspectives would be vital to ensure that the development of \texttt{TAI} aligns with human values, mitigates risks, and maximizes its benefits for all.

Approaching and understanding AI through the lenses of Ethics is paramount in ensuring its responsible and beneficial development. The next section helps us understand the relevance of wrapping AI in an ethical framework, and how to deal with the possible arrival of \texttt{TAI}.
\section{The Relevance of Ethical Perspectives in AI}\label{ethics}

The way humans' ethics perceives AI is of paramount importance in its path toward achieving \texttt{TAI}, becoming into a disruptive and enabling technology in our civilization. This perception affects not only its adoption and regulation, but also the development of applications in key areas such as medicine, economics, and decision-making. Today we live in a world dominated by the use of technology, and in particular the application of AI everywhere. There are several philosophical perspectives on AI that we should consider to really understand how and why we have reached such a point; each one providing unique insights and raising distinct ethical, metaphysical, and epistemological questions (Table \ref{tab:ethics_persp}). All of them conform to the current AI understanding and the global view, essential to comprehend the impact and consequences of our actions and decisions, our interests, our morals, and our future as a civilization. The purpose of this section is far from being an exhaustive analysis of the different philosophical theories, but to indicate how the most widespread ones have had, have, and will have, a great influence in explaining how each of us perceives AI, and in what form \texttt{TAI} could be adopted (or not):

\begin{table}[h!]
\centering
\renewcommand{\arraystretch}{1.5}
\begin{tabularx}{\textwidth}{@{}>{\centering\arraybackslash}p{3.5cm}X@{}}
\toprule
\makecell{\textbf{Ethical}\\ \textbf{Perspective}} & \makecell{\textbf{Key Contribution}\\ \textbf{and Application in AI}} \\ 
\midrule
\texttt{Utilitarianism} & It maximizes overall well-being and minimizes harm. For example, designing AI systems that optimize collective benefits (e.g., reducing traffic accidents) while protecting individual rights. \\ 
\midrule
\texttt{Deontology} & It establishes universal rules based on rights and obligations. For instance, creating AI that respects fundamental norms (e.g., non-discrimination) but allows reasonable exceptions (e.g., in emergencies). \\ 
\midrule
\texttt{Virtue Ethics} & It promotes values such as justice, compassion, and prudence in decision-making. For example, developing AI systems that reflect shared ethical values, such as fairness and transparency in decision-making algorithms. \\ 
\midrule
\texttt{Ethics of Care} & It prioritizes empathy, human relationships, and support for the most vulnerable. For instance, ensuring AI benefits vulnerable populations, such as enhancing accessibility for individuals with disabilities or protecting marginalized groups. \\ 
\midrule
\texttt{Contractualism} & It is based on mutual agreements acceptable to all affected parties. This includes implementing standards for AI development that balance innovation and equity, respecting fundamental rights while maximizing societal benefits. \\ 
\bottomrule
\end{tabularx}
\vspace{2mm}
\caption{Overview of some representative ethical perspectives in AI.}
\label{tab:ethics_persp}
\end{table}

\paragraph{ $\bullet$ Utilitarianism} From an utilitarian perspective, the development and application of AI should focus on maximizing overall happiness and minimizing suffering. Utilitarianism, as a consequentialist theory, evaluates the morality of actions based on their outcomes, and in the context of AI, this implies that AI systems should be designed and employed to generate the greatest possible benefit for the largest number of people. For instance, AI could be used in healthcare to optimize diagnoses, improve treatment outcomes, and allocate resources efficiently, thereby improving the well-being of entire populations. Similarly, AI can enhance societal welfare by automating dangerous or tedious tasks, reducing risks to human workers, and minimizing suffering. However, utilitarianism also faces ethical challenges in the context of AI, particularly when it comes to privacy and individual rights. Utilitarian principles could justify actions like the use of personal data without consent if doing so results in a greater good for the majority, which raises concerns about fairness and exploitation. Therefore, while utilitarianism emphasizes the benefits of AI for the majority, it must also carefully balance these benefits against potential harms to individuals or minority groups \citep{cvik2022categorization}. 

\paragraph{$\bullet$ Deontological Ethics} It focuses on the adherence to moral duties and principles, regardless of the consequences. In the context of AI, deontological ethics emphasizes the importance of designing AI systems that respect fundamental rights and adhere to strict ethical standards, regardless of the outcomes those systems may produce \citep{mougan2023kantian}. A deontological approach to AI prioritizes respect for individual privacy, autonomy, and dignity, ensuring that AI technologies do not violate moral rules such as the right to personal data protection or the right to be treated equally. For instance, deontological ethics demands that AI systems used in hiring or law enforcement are free from bias and discrimination, respecting each person’s inherent value. Transparency and accountability are also critical in a deontological framework, where AI systems must provide clear justifications for their decisions and allow for human oversight. This perspective often clashes with more consequentialist (e.g., utilitarianism) approaches because it upholds the primacy of ethical rules, even when doing so may limit the overall benefits AI could provide. Deontological ethics in AI emphasizes the need for moral consistency, fairness, and respect for universal principles, which can conflict with purely outcome-based strategies.

\paragraph{$\bullet$ Virtue Ethics (the Aristotelian View)} It addresses the development of moral character and the cultivation of virtues such as wisdom, courage, and justice. To AI, this approach encourages the design and application of AI systems in ways that promote human flourishing (\textit{eudaimonia}) and support the development of virtuous behaviors. Rather than focusing solely on rules or outcomes, virtue ethics in AI looks at how these technologies can enhance the moral and intellectual development of individuals and society. For example, educational AI systems could be designed not only to improve academic performance but also to encourage traits such as perseverance, curiosity, and ethical reasoning. Similarly, AI applications in healthcare could promote holistic well-being by supporting patients in building habits that enhance both their physical and mental health. Virtue ethics also places a strong emphasis on the communal aspect of human life, suggesting that AI should be designed to foster cooperation, empathy, and social harmony. However, a significant challenge for applying virtue ethics to AI lies in its lack of specific guidelines for addressing complex moral dilemmas \citep{Hagendorff_2022}, making it challenging to implement in technical systems that depend on clear, explicit instructions.

\paragraph{$\bullet$ Contractualism} John Rawls focused on the idea that moral principles are those that individuals would agree under fair conditions \citep{rawls2017theory}. In the context of AI, this theory suggests that the ethical development and application of AI systems must be grounded in rules and practices that all rational individuals would accept as just and fair. This perspective is particularly relevant in areas like algorithmic fairness and justice, where AI systems used in hiring, criminal justice, or credit scoring, must be designed to prevent discrimination and promote equality. Under a contractualist framework, AI systems should be transparent, accountable, and designed with the input of all stakeholders to ensure that their decisions are perceived as legitimate and just. Additionally, Rawls’ concept of the ``veil of ignorance'' \citep{rawls2017theory}, where decision-makers are asked to design systems without knowing their own position in society, could guide the creation of AI technologies that prioritize the needs of the most disadvantaged. This approach emphasizes the social contract and the need for fairness in the distribution of AI's benefits and burdens, but it can be criticized for being too idealistic or difficult to implement in practice, especially in contexts where societal agreements are not easily reached.

\paragraph{$\bullet$ Ethics of Care} It emphasizes relationships, empathy, and responsibility toward others, offering a distinct approach to AI that focuses on using technology to foster care and support for individuals, particularly the vulnerable \citep{held2006ethics}. In AI applications, this theory advocates for systems that prioritize human well-being, emotional support, and relational bonds, rather than merely efficiency or profit. For example, AI systems in healthcare could be designed to assist caregivers in providing personalized, compassionate care for the elderly, children, or individuals with disabilities. AI could also be used in social robotics or mental health services to offer emotional support, enhancing the emotional and psychological well-being of users. The ethics of care stresses the importance of context and the specific needs of individuals, which implies that AI systems should be adaptable to different caregiving situations and should consider the emotional and social impact of their interactions with humans. However, a potential criticism of this approach is that it may not scale easily to larger, impersonal systems or more complex societal issues, and it might struggle to provide clear ethical guidelines for AI decision-making in areas that extend beyond personal relationships or caregiving contexts.

\subsection{Challenges in Translating Ethical Perspectives to AI}

Articulating ethics in AI is inherently challenging due to the vast diversity of beliefs, cultural values, and moral frameworks that exist globally. Ethical principles are deeply rooted in social, historical and religious contexts, which vary significantly across regions and communities. For instance, a principle deemed ethically sound in one culture may conflict with the values of another. This pluralism complicates the process of defining universal ethical standards for AI systems, especially those designed for deployment on a global scale. Moreover, philosophical traditions differ in their approaches to morality. Western ethical frameworks, such as utilitarianism and deontology, often emphasize individual rights and rational decision-making, whereas non-Western perspectives, such as Confucianism or Ubuntu, may prioritize relational harmony and communal well-being. These differences create tension in determining which ethical principles should guide AI development, usage, and governance. 

Translating ethics into rules is the most practical and accessible way to apply ethical principles \citep{Hagendorff_2022}, particularly in AI, offering clarity, consistency, and scalability, which are essential for integrating ethical considerations into decision-making processes. By converting abstract ethical principles into explicit rules, systems can operate autonomously while adhering to predefined moral standards. However, translating ethical perspectives into concrete rules for implementation in AI depends on the structure and principles of each framework. Some ethical theories, due to their normative clarity and formal structure, lend themselves more readily to codification:
\begin{itemize}
\setlength{\itemsep}{0.8em}
    \item Deontological ethics is inherently rule-based, providing clear guidelines such as ``do not lie'' or ``or not harm'', which can be directly encoded into algorithms as constraints or rules. However, while clear, deontological rules can be rigid and may struggle to adapt to complex scenarios where duties conflict, such as when respecting privacy may impede public safety.
    \item Utilitarian ethics is mathematically compatible with algorithmic systems, as AI can optimize predefined utility functions to maximize benefits or minimize harm. However, it requires quantifying well-being, which is often subjective and culturally dependent. Additionally, it may sacrifice individual rights for the collective good, leading to ethical dilemmas.
    \item Contractualism lends itself well to regulatory frameworks and compliance models in AI, where systems are designed to adhere to established policies or social contracts. However, achieving global consensus on ethical principles is difficult particularly in culturally diverse contexts, as mentioned before.
\end{itemize}

Conversely, others are inherently interpretative or context-dependent, making them challenging to operationalize:
\begin{itemize}
\setlength{\itemsep}{0.8em}
    \item The concepts of virtue ethics are abstract and context-dependent, lacking clear rules for specific actions. Translating virtues into operational rules for AI is challenging because virtues often require subjective judgment and adaptability. While an AI system can mimic virtuous behavior, truly embodying virtues such as empathy or fairness in a human sense may exceed current technological capabilities.
    \item Unlike rule-based systems, ethics of care relies on understanding individual needs and fostering connections, which are inherently dynamic and situational. AI systems, which operate on generalized models, struggle to capture the nuances of relational ethics. Implementing ethics of care requires AI systems capable of deep contextual understanding and emotional intelligence, both of which are still nascent fields.
\end{itemize}

\begin{figure}[H]
    \centering
    \includegraphics[width=0.65\linewidth]{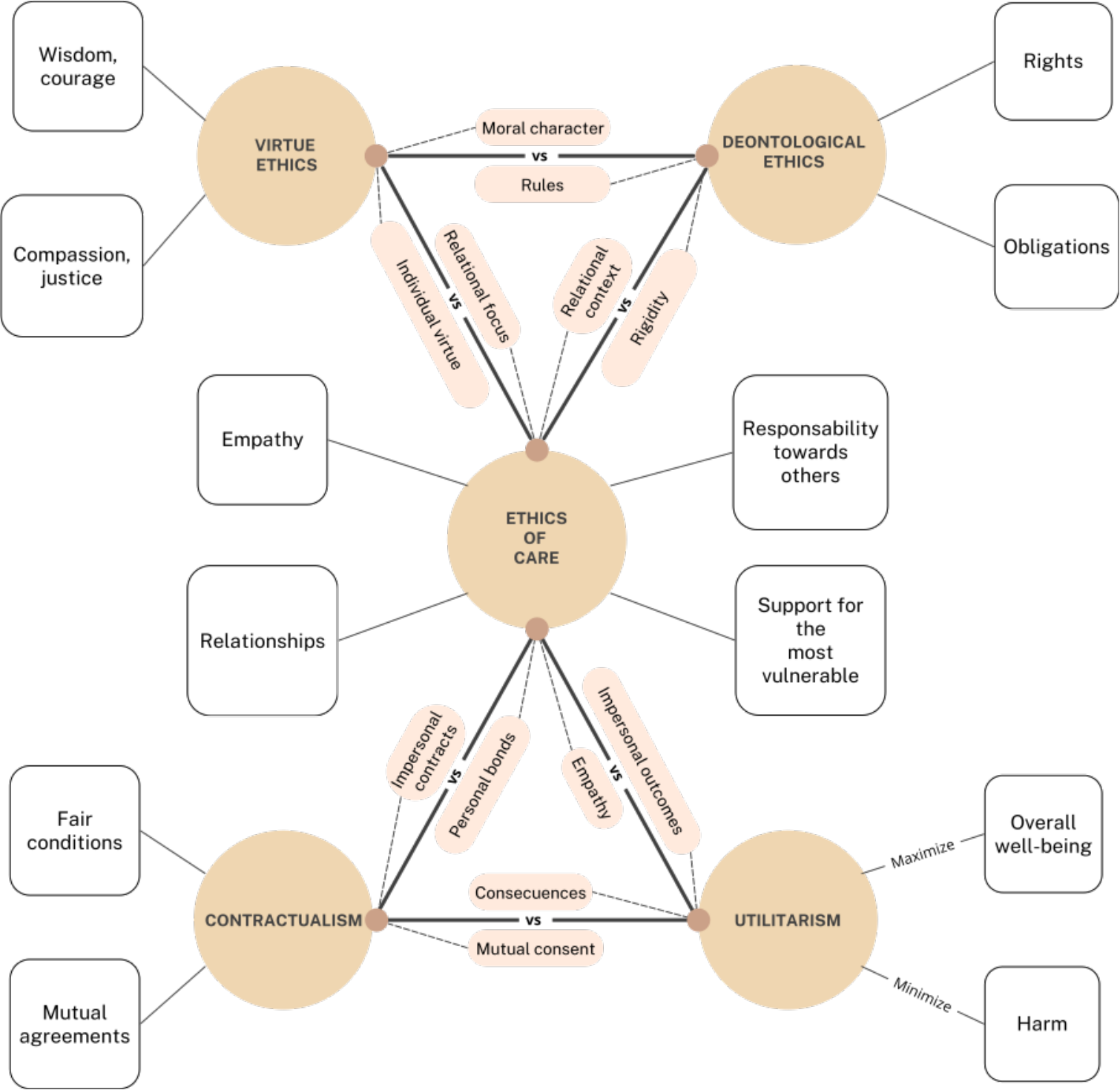}
    \caption{Conflicts and principles of the main ethical perspectives in AI.}
    \label{fig:conflicts}
\end{figure}

Finally, many of the major ethical perspectives, such as utilitarianism, deontologism, or virtue ethics often seem to conflict with each other because of their different approaches to what constitutes morally a right action (see Figure \ref{fig:conflicts}). Utilitarianism and deontologism often clash when rules or duties (deontological principles) contradict actions that could maximize overall good (utilitarian outcomes). Likewise, virtue ethics may not always provide clear guidelines for action, especially when outcomes are ambiguous.  

We then found some factors affecting the translatability of ethical perspectives to AI. The clearer and more prescriptive the ethical framework, the easier it is to translate into rules; deontology and utilitarianism offer explicit guidelines, while virtue ethics and ethics of care rely on interpretative judgment (clarity of norms). Ethical perspectives that require context-specific reasoning, such as care ethics, are more challenging to generalize into fixed rules suitable for AI (context dependency). Ethical frameworks differ in their ability to handle conflicts; deontology struggles with competing duties, while utilitarianism risks controversial trade-offs, complicating their practical application (conflict resolution). The latter is particularly relevant in ethical frameworks, especially when applied to AI, because ethical dilemmas often arise from competing values, duties, or interests. The ability to navigate and resolve these conflicts is essential to ensure that ethical principles are not only clearly defined but also actionable in practice. This underscores the need for their reconciliation, which is addressed in the following section.

\subsection{Reconciliation between Ethical Perspectives}\label{ssec:reconciling}

We have seen how difficult is to bring some of the ethical perspectives into practice. However, there are alternatives to reconcile some of these perspectives into a more complementary and balanced ethical framework, capable of more robustly addressing the ethical challenges posed by AI, and which could facilitate its implementation. Next we present some examples:

\paragraph{$\bullet$ Virtue Ethics Combined with Rule-based Approaches} Virtue ethics focuses on the moral character and virtues of the individuals or systems involved, such as fairness, responsibility, and transparency. To reconcile this with deontological and consequentialist approaches, AI systems can be designed not only to follow rules but also to foster virtuous traits. For example, AI decision-making systems could be structured to prioritize transparency and fairness, while also embracing utilitarian principles when maximizing societal benefit (e.g., in healthcare or resource distribution). A practical application could involve combining virtue ethics to promote traits like trustworthiness and compassion while using deontological rules to set limits (e.g., ensuring privacy), and utilitarianism to evaluate the overall benefits.
    
\paragraph{$\bullet$ Ethics of Care and Utilitarianism} Ethics of care emphasizes relational and contextual ethics, focusing on the well-being of specific individuals or groups. It can complement utilitarianism by addressing the blind spots in general welfare maximization, particularly for vulnerable populations. AI systems, under this re\-conciliation, could be designed to prioritize care for vulnerable individuals (e.g., in eldercare or personalized healthcare), while still optimizing overall social welfare. In designing AI for social services, algorithms could balance caring for individual clients with the utilitarian goal of maximizing overall service efficiency.
    
\paragraph{$\bullet$ Utilitarianism with Deontological Safeguards} Utilitarianism often prioritizes outcomes, sometimes at the expense of inviolable rights. To reconcile this, deontological principles can provide essential safeguards. For instance, AI-driven systems could maximize beneficial outcomes (such as reducing traffic accidents) but must operate within strict rules that protect individual rights, like privacy or non-discrimination. In autonomous driving, the system could be designed to minimize accidents (a utilitarian goal) but also ensure strict adherence to human rights laws regarding data privacy and safety standards.
    
\paragraph{$\bullet$ Through Contractualism} Based on the principles of fairness and mutual agreement, contractualism can serve as a foundational framework to reconcile other ethical theories. By developing AI under a social contract where stakeholders agree on fair terms, conflicting ethical perspectives, such as utilitarianism and deontologism, can be balanced. For example, contractualism can help define when to prioritize individual rights (deontological limits) and when to maximize the common good (utilitarian concerns). Then, Rawls' ``veil of ignorance'' \citep{rawls2017theory} can be applied to ensure that AI technologies are developed and deployed in fair ways, especially for the most disadvantaged. It allows the balancing of utilitarian outcomes (benefits for the majority) with deontological commitments to protect fundamental rights.

\vspace{\baselineskip}

\begin{figure}[H]
    \centering
    \includegraphics[width=0.8\linewidth]{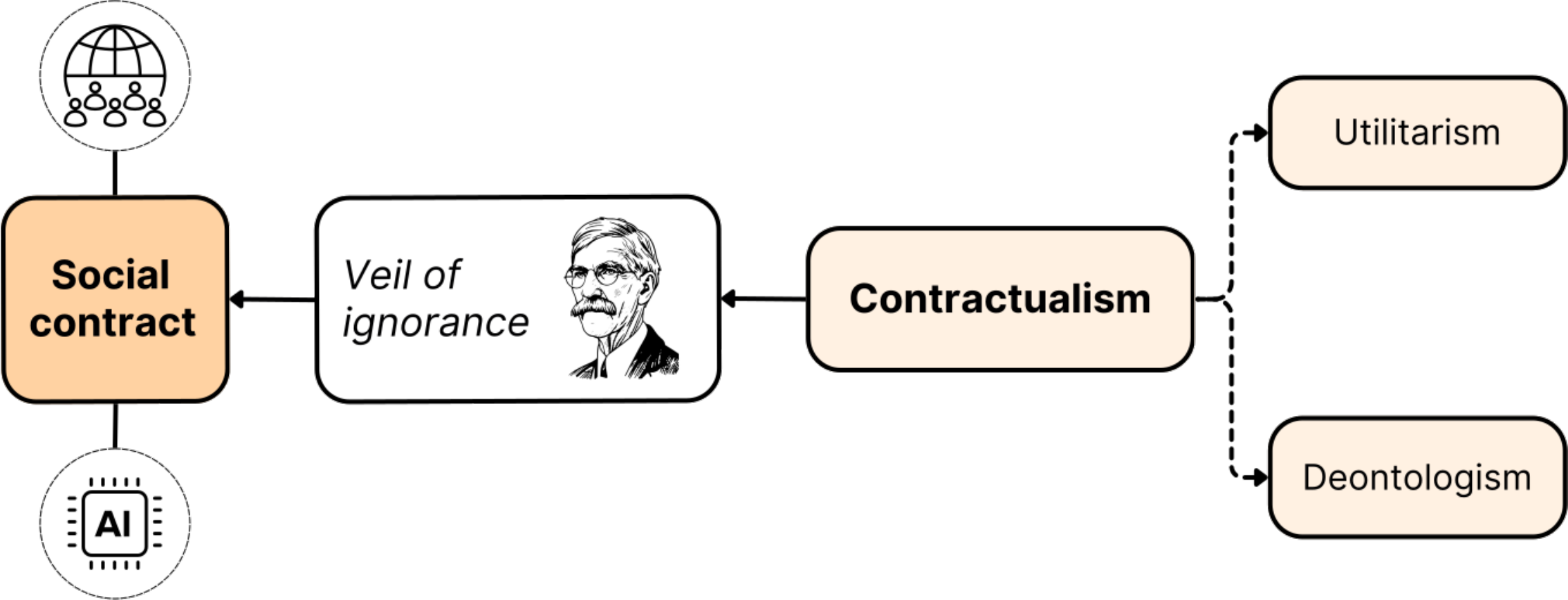}
    \caption{Example of contractualism as a reconciliatory framework for AI.}
    \label{fig:contract_recon}
\end{figure}

Contractualism as a reconciliatory framework is especially interesting, as it seeks a balanced and equitable approach. As shown in Figure \ref{fig:contract_recon}, contractualism uses the concept of a ``social contract'' to integrate other ethical theories, which facilitates consensus among various stakeholders. Its focus on rational consensus and fair principles can be regarded as a way to align AI regulations with democratic norms. This approach is particularly useful for environments where multiple stakeholders must agree on ethical and normative principles (e.g., regulatory frameworks such as the \textit{EU AI Act}\footnote{\label{eu_act}\url{https://artificialintelligenceact.eu/} [Accessed on December 10th, 2024].}). Being rule-based, contractualism can be more easily adaptable to legal and political regulations, and it allows for a nuanced assessment of different levels of risk while ensuring that both individual rights and societal well-being are taken into account. It is successful in creating inclusive and balanced policies, but its application in highly dynamic systems, such as rapidly evolving AI systems, can be complex due to the difficulty of anticipating all circumstances in which AI-based systems could be used over time.

In the next section, the aforementioned \textit{EU AI Act} serves as a practical example of reconciliation between different ethical perspectives.

\subsection{Unpacking the Ethical Perspectives of the \textit{EU AI Act}}\label{ai_act_ethics}

The AI Act has been developed as a comprehensive regulatory response to the rapid advances and increasing deployment of AI technologies. Its main objective is to ensure that the use of AI in Europe is safe, ethical and respectful of fundamental rights, while fostering innovation and protecting the European single market \citep{diaz2023connecting}. But its application is not exempt of its own problems \citep{novelli2024taking}. We must look at the Act through this lens, that of a single ethical perspective capable of reconciling with others and possessing an inherently integrative capacity, thus being ethically grounded in a combination of theories, reflecting a pluralistic ethical framework. The Act can be seen as a formalization of a social contract between governments, developers, businesses, and the public, establishing clear rules to regulate the use of AI in a way that is justifiable to all parties involved, and that reconciles several ethical perspectives:

\paragraph{$\bullet$ Contractualism} The Act's emphasis on transparency, accountability, and the protection of fundamental rights aligns with the contractualist idea that moral actions should be acceptable to all rational agents. By requiring that AI systems be designed and deployed in ways that respect human dignity, the Act can be understood as a set of agreed-upon norms to prevent harm and promote fairness. It aims to ensure that no individual is unfairly disadvantaged by AI systems, particularly in high-risk areas such as healthcare, employment, and law enforcement. Additionally, the risk-based framework can be seen as a reflection of contractualist principles, as it distinguishes between acceptable and unacceptable levels of risk, ensuring that individuals are not exposed to AI systems that pose unreasonable risks to their well-being.

\paragraph{$\bullet$ Deontologism} The emphasis of the Act on human dignity, transparency, and accountability can be viewed through a deontological lens as an affirmation of the moral duty to respect individuals as ends in themselves, not merely as means to technological advancement or economic gain. The requirements of the Act for informed consent and explainability in high-risk AI systems, such as biometric identification or AI used in law enforcement, reflect a deontological commitment to respecting individuals’ rights to autonomy and informed decision-making. It recognizes that humans should not be treated as passive subjects under AI systems without understanding how these systems impact them or having the ability to challenge AI-driven decisions. The duty to avoid harm, central to deontological ethics, is also mirrored in the safety requirements for AI systems. The Act imposes strict guidelines to prevent foreseeable risks, ensuring that developers and deployers of AI systems act under their moral responsibilities to protect individuals from harm, even if doing so may limit certain technological capabilities or innovations.

\paragraph{$\bullet$ Utilitarianism} The Act can be interpreted as an attempt to balance the benefits and risks of AI by creating a regulatory framework that maximizes the positive impact of AI technologies while minimizing potential harm to individuals and society. The risk-based approach of the Act to AI governance reflects a utilitarian concern with ensuring that AI systems provide net benefits to society. High-risk AI systems are subject to stricter regulations to ensure that they do not cause disproportionate harm compared to their potential benefits. For instance, AI used in critical infrastructure or law enforcement must undergo rigorous scrutiny to ensure that it enhances societal well-being without infringing on individuals' rights or safety. Additionally, the focus on innovation and the promotion of AI technologies in sectors like healthcare, where they can lead to improved outcomes for many, aligns with the utilitarian goal of promoting the greatest good for the greatest number. However, utilitarianism also raises concerns about the trade-offs involved: the Act must ensure that maximizing societal benefits does not unjustly sacrifice the rights of vulnerable individuals or groups, especially in cases where AI systems may inadvertently perpetuate biases or inequalities.

\paragraph{$\bullet$ Virtue Ethics} The Act can be understood as encouraging the development of virtuous practices in AI development and deployment by promoting traits like responsibility, transparency, fairness, and trustworthiness. By requiring developers and users of AI to be accountable for the systems they create and deploy, the Act fosters a sense of moral responsibility and ethical reflection. Virtuous AI developers are expected to take into account the social and moral implications of their technologies, ensuring that their work contributes to the common good and does not cause harm. The focus of the Act on trustworthiness aligns with the virtue of integrity. Trustworthy AI systems are those that are reliable, transparent, and fair. The regulatory requirements for algorithmic transparency and explainability aim to ensure that AI systems are designed and used in ways that reflect the virtues of honesty and respect for others, promoting public confidence in the ethical use of AI.

\paragraph{$\bullet$ Rawlsian Justice} The Act can also be interpreted through the lens of justice as fairness. According to Rawls, a just society is one that ensures equal rights and fair opportunities for all individuals, particularly the most vulnerable. The focus of the Act is on preventing discrimination and ensuring that AI systems are deployed in ways that are fair and just, aligning with Rawlsian principles. The commitment of the Act to preventing bias and ensuring fairness in AI systems (particularly in areas such as employment, credit scoring, and law enforcement) reflects the Rawlsian concern for protecting those who might be most negatively affected by biased or unfair AI decisions. The Act ensures that AI systems do not exacerbate existing inequalities and that they provide fair opportunities and outcomes for all individuals, regardless of their socio-economic status, gender, race, or other characteristics. Additionally, the transparency and accountability mechanisms built into the Act ensure that decisions made by AI systems can be scrutinized and challenged, allowing individuals to appeal AI-driven decisions that may affect their rights or opportunities. This promotes procedural justice, ensuring that the processes governing AI systems are fair and inclusive.

\vspace{\baselineskip}

After presenting some of the most relevant ethical perspectives on AI and their consideration in the \textit{EU AI Act}, the next chapter explores how AI has come to play such a significant role in our lives, as it has in the past.
\chapter{From AI Today to the Challenges of TAI}\label{part_2}
\section{The Omnipresence of AI Today}\label{here_again}

AI is not a field that has recently emerged with the appearance of generative AI and the next-generation chatbots. AI as a field of study has been with us since the 1950s \citep{mccarthy2006proposal}, and has always raised expectations that have so far been difficult to meet. The truth is that since ancient times, we can find references and ideas about the creation of beings with human characteristics, automata, or mechanical devices with mathematical and logical foundations, up to the appearance of the Turing machine \citep{turing1936computable} as a prelude to this new field of study.

\subsection{A Time in the Shadows}

\begin{figure}[H]
    \centering
    \includegraphics[width=0.9\linewidth]{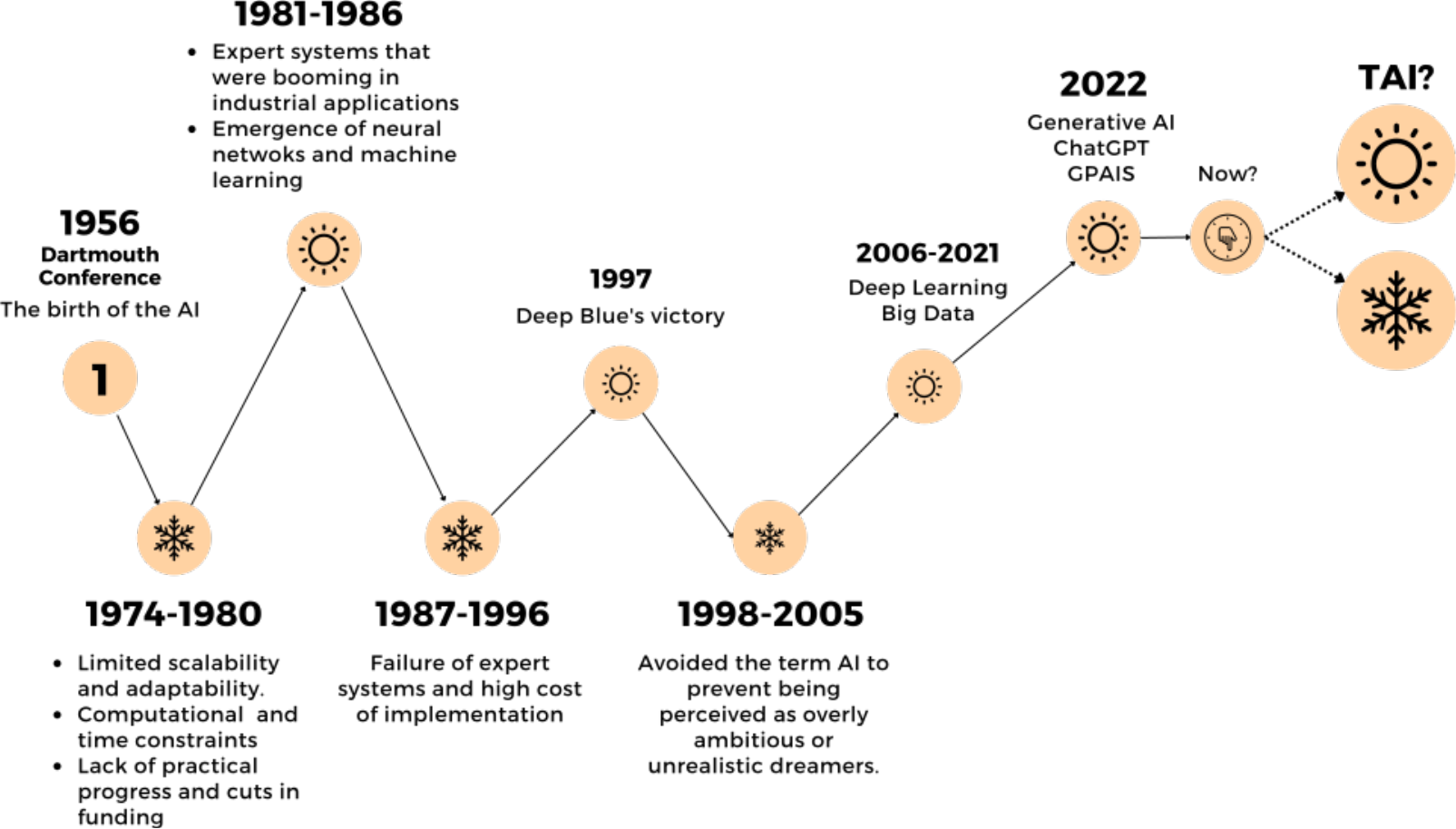}
    \caption{A historical overview of AI cycles, and the opportunity to become a truly transformative turning point in the near future.}
    \label{fig:AI_time_line}
\end{figure}

Since its inception, the field of AI has experienced several cycles of optimism and stagnation, often referred to as ``AI summers'' and ``AI winters'' (see Figure \ref{fig:AI_time_line}). These periods of intense research enthusiasm and investment have alternated with phases of disillusionment and reduced funding, as early technological promises failed to materialize. In the mid-20th century, initial advancements in symbolic reasoning and rule-based systems sparked the first wave of excitement. However, the limited scalability and adaptability of these systems, coupled with the computational constraints of the time, led to the first major AI winter in the 1970s. The emergence of machine learning, particularly neural networks, reinvigorated the field in the 1980s, ushering in another ``AI summer''. Yet, the challenges of training deep networks and the lack of large-scale data again tempered expectations, resulting in a second AI winter in the late 1980s and early 1990s. In recent years, however, AI has undergone a profound resurgence, driven by the confluence of increased computational power, vast datasets, and advancements in algorithmic techniques (e.g., deep learning). Whether its role in the future could be key, or purely speculative, forming part of yet another ``technology bubble'', is still a matter of debate \citep{floridi2024ai}.

Particularly, the advent of modern generative AI models has marked a transformative phase in AI development. These models, capable of generating human-like text, code, and even creative content, have demonstrated a level of utility and sophistication that is unparalleled in previous decades. The case of \textit{ChatGPT}\footnote{\url{https://openai.com/index/chatgpt/} [Accessed on December 10th, 2024].} is particularly interesting; this tool revolutionized the accessibility of AI in November 2022, effectively democratizing a technology that was once confined to research labs and specialized industries. By providing a user-friendly interface and powerful natural language processing capabilities, \textit{ChatGPT} enabled e.g., individuals, businesses, or educators to interact with AI in ways that were previously unattainable. This democratization has transformed AI into an everyday tool, empowering users to generate creative content, enhance learning, and streamline workflows without requiring technical expertise. As a result, \textit{ChatGPT} not only broadened public engagement with AI but also redefined its role in society, making advanced AI capabilities accessible to millions of users worldwide.

For the first time, AI is perceived not merely as a tool for automating tasks but as a genuinely transformative technology with wide-ranging implications across industries and scientific domains. Furthermore, this progress has led some researchers to consider, with newfound seriousness according to some, the possibility of achieving \texttt{BCAI} through a system capable of performing any intellectual task that a human can \citep{morris2023levels}. Efforts to develop best practices, standards, and regulations for such systems are also emerging \citep{schuett2023towards}. While \texttt{BCAI} seems to remain a distant goal, the vibrant momentum of AI technologies suggests that such ambitions might no longer be confined to the realm of speculative fiction, but rather a plausible, albeit long-term, scientific endeavor. 

\subsection{AI Until Now}\label{until_now}

At present, the field of AI is dominated by advancements in \textit{narrow AI} or \textit{weak AI}, which refers to systems designed to perform specific tasks with high efficiency and accuracy. These systems, while highly specialized, do not possess a general purpose or an understanding of the tasks they perform in a broader sense. Prominent among these are deep learning models, machine learning algorithms, and most notably, generative AI, exemplified by Large Language Models (LLMs). These models have the capacity to generate human-like text, create content, and even assist in coding and problem-solving. However, they remain task-specific and lack the general reasoning capabilities of human intelligence. Nowadays we can find the so-called \emph{General Purpose Artificial Intelligence Systems} (GPAIS) \citep{triguero2024general}, referring to AI that is capable of being autonomous to solve more than one task, and of generalizing to unseen tasks. Nevertheless, AI is still highly dependent on humans, and still lacks abilities such as complex reasoning o truly understand what it is doing. This is why the AI we have today is still confined to the \textit{narrow AI} category. This statement is not without debate lately, as while some believe that some capabilities demonstrated by generative AI such as \textit{ChatGPT} can be considered as \texttt{BCAI} (concretely as emerging AGI \citep{morris2023levels}) or the future of AI \citep{sejnowski2024chatgpt}, while many others still consider it \textit{narrow AI} since it is still limited to a single text-based chat task. 

Despite their limitations, \textit{narrow AI} systems are currently driving profound transformations across multiple sectors. without making as much noise or achieving the advanced capabilities of \texttt{BCAI}. In business, AI is revolutionizing industries by optimizing operations, enhancing customer service through automation and personalization, and enabling predictive analytics that improve decision-making. Moreover, generative AI is reshaping creative industries by producing art, music, and written content at scales and speeds previously unimaginable. Socially, AI is altering how individuals interact with technology, from virtual assistants becoming integral parts of daily life \citep{gabriel2024ethics} to AI-driven systems influencing social media algorithms and content consumption patterns. This ongoing revolution is characterized by increased efficiency, cost reduction, and the democratization of specialized knowledge through AI tools accessible to a broad user base. Yet, the transition toward \texttt{TAI} might hinge on overcoming the current limitations of \textit{narrow AI}, or advancing \texttt{BCAI}. While \texttt{BCAI} remains speculative, the societal and business impacts of current AI technologies are undeniable, signaling that we could be on the cusp of unprecedented technological and social change. 

Nonetheless, if we really expect AI to one day be truly transformative and catapult our civilization to another stage, we must do more things, yet differently. The achievement of \texttt{TAI} depends mostly on what role we give to AI in our society and civilization, and on our ability to overcome the stones we encounter along the way. AI today lacks a clearly defined role in our society, having expanded rapidly across various domains without sufficient reflection on its purpose, justification, and appropriate applications. While its potential is vast, the deployment of AI often appears driven by technological breakthroughs and economical interests rather than carefully considered needs or goals. Questions such as why we need AI in specific contexts, under what circumstances it should be applied, and how it should be integrated, remain largely unanswered or at least without consensus. This uncoordinated proliferation risks misaligned priorities, wasted resources, and unintended consequences, as AI systems are deployed without well-defined frameworks to guide their development and use. To harness the transformative potential of AI effectively, it is imperative to define its role more coordinately, ensuring that it serves well-defined societal, ethical, and practical objectives rather than existing as a ubiquitous but aimless innovation.

\subsection{The AI Role in Our Civilization}\label{AI_role}

For the first time, our society sees AI as a leading force capable of guiding and catapulting our civilization to the next stage. It is now that we can see certain parallels with the \textit{Plato's Allegory of the Cave} \citep{plato_republic_jowett}, using it as a metaphorical lens. 

\begin{quote}
    ``And if someone dragged him away from there by force, up the rough ascent, the steep way up, and never stopped until he could see the light of the sun...''
    \\
    --- \textit{Allegory of the Cave, Plato}
\end{quote}

Under such parallelism, AI would exercise its role of ``mediator between shadows and reality''. In the allegory, the prisoners in the cave perceive shadows on the wall as reality because they lack access to the world beyond their immediate senses. Similarly, AI systems operate within a framework of ``shadows'' (the data they are trained on, which represents a limited and curated subset of reality). \texttt{TAI}, however, could act as a mediator, analyzing patterns in these shadows and uncovering deeper truths that might remain inaccessible to human cognition alone. This mediation aligns AI with the role of the philosopher in such an allegory, seeking to guide humanity toward a clearer understanding of the world. AI would also exercise its role as a ``chain breaker''. The chains in the cave symbolize the cognitive and perceptual limitations of human beings. AI would have the potential to break these chains by augmenting human capabilities, processing vast amounts of data, and offering insights into complex systems. Just as the philosopher ventures outside the cave to experience the light of the sun (truth), \texttt{TAI} could help humanity transcend its limitations and access realms of knowledge previously beyond its grasp. 

However, there is another point of view that contrasts with the above. The allegory also warns of the deceptive nature of shadows, which can be manipulated by those controlling the light source. Similarly, AI systems can reinforce biases, perpetuate misinformation, or present an illusion of truth if their training data or algorithms are flawed. This raises ethical concerns about who controls AI systems and how they are designed, echoing the allegory's caution against mistaking shadows for reality.

\begin{figure}[ht]
    \centering
    \includegraphics[width=0.9\linewidth]{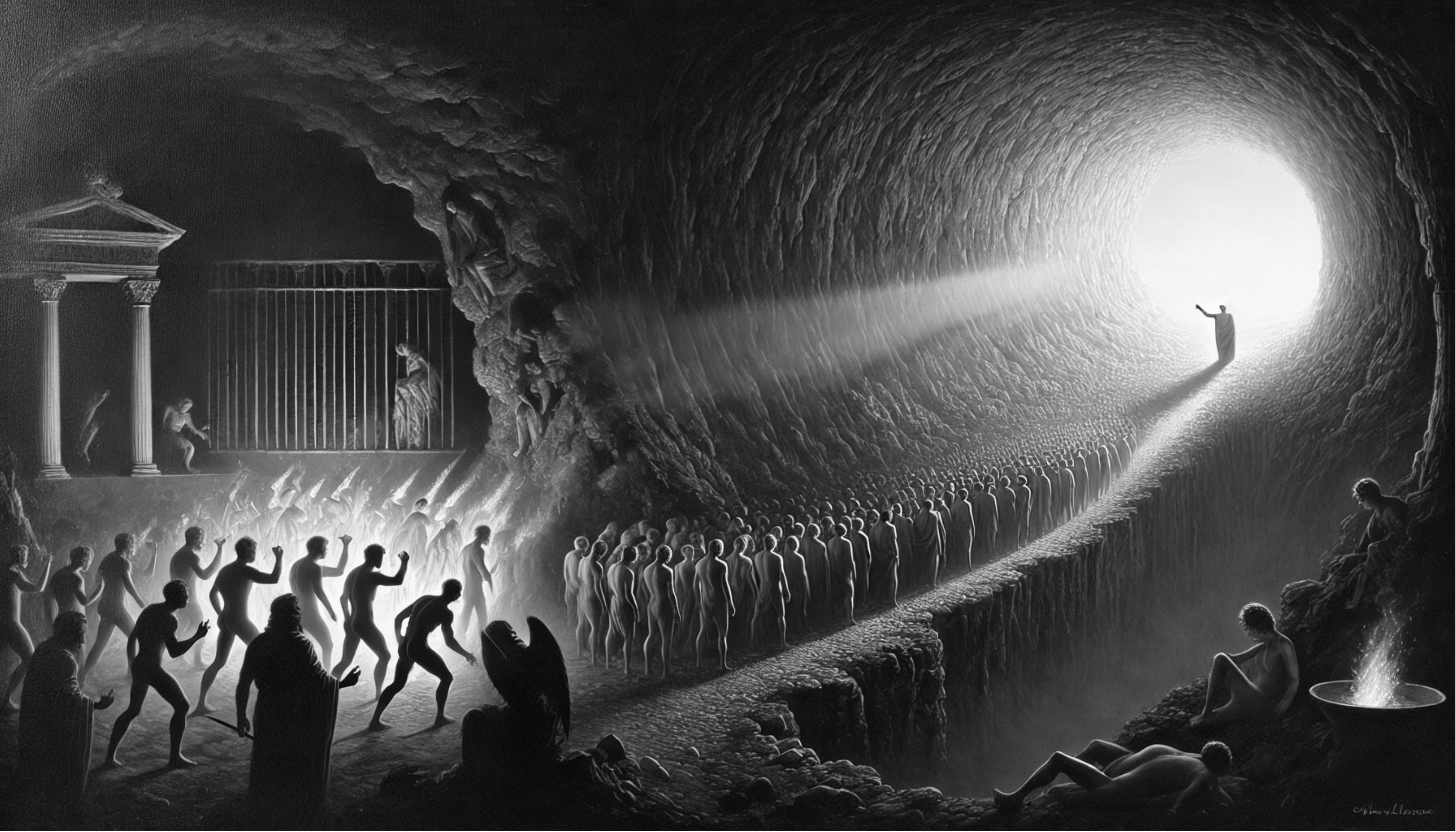}
    \caption[The parallelism with the Plato's Allegory of the Cave]{This illustration depicts the parallelism explained in this section. Image generated with the assistance of \textit{ChatGPT}, a language model developed by OpenAI.}
    \label{fig:cave}
\end{figure}

Many hopes are pinned on truly transformative AI, and there are many hurdles to get there. At the moment, today's AI is permeating almost every layer of our society, and is taking on a number of roles in many facets of our lives. The role of AI in contemporary civilization can be analyzed from two contrasting perspectives: one in which we increasingly delegate many of our processes and decisions to AI, and another one where AI serves as a collaborator to augment human capabilities. These two approaches reflect different attitudes toward the integration of AI in society, and have profound implications for \texttt{TAI}, the future of human agency, autonomy, and progress:

\paragraph{$\bullet$ Delegating Processes to AI: A Path of Automation and Efficiency} The potential of AI is harnessed to take over a substantial portion of decision-making and operational processes across various sectors. This vision is often characterized by a focus on automation, where AI systems assume tasks traditionally performed by humans, optimizing for speed, accuracy, and cost-effectiveness. AI-driven automation has the capacity of significantly enhancing productivity in industries such as manufacturing, logistics, finance, and healthcare. By automating repetitive and data-intensive tasks, businesses can achieve higher efficiency levels and reduce human error. This can lead to lower costs, faster production cycles, and the ability to scale operations globally. This perspective envisions a world where humans are less involved in routine decision-making processes, and many aspects of everyday life are managed by intelligent systems. In such a scenario, AI systems may take over processes like driving, content creation, or even aspects of governance, such as resource allocation or law enforcement. By reducing human intervention, these AI systems can potentially minimize biases and inefficiencies inherent in human decision-making. 

However, this approach comes with significant risks. By increasingly granting control to AI, there is a concern that human agency may diminish, leading to a society where decisions could be opaque and determined by algorithms that might not fully align with human values. Additionally, this could concentrate power in the hands of those who control advanced AI systems \citep{coeckelbergh2024artificial}, potentially creating new forms of inequality. The risk of dependency on AI systems also gives rise to safety and security issues where critical functions of society could falter if these systems encounter unexpected failures or adversarial attacks.

\paragraph{$\bullet$ AI as a Collaborative Partner: Enhancing Human Potential} This envisions AI as a collaborator that works alongside humans to enhance our cognitive and physical capabilities rather than replace them. This approach emphasizes the synergistic potential of AI to complement human strengths and compensate for human limitations, fostering a more holistic development of society. When used as a collaborative tool, AI has the potential to augment human creativity, insight, and decision-making processes. Collaborative AI can be designed to offload routine cognitive tasks, allowing humans to focus on more complex, creative, and strategic aspects of their work. This empowerment would allow humans to tackle challenges that require empathy, ethical consideration, and adaptability, qualities that are difficult to encode and thus remain beyond the reach of current AI capabilities. This perspective requires a commitment to human-centric AI design, where the primary goal is to ensure that AI aligns with human values and interests. This involves creating AI systems that are transparent, explainable, and accountable, enabling humans to understand and guide AI decisions. Additionally, this approach emphasizes the importance of retaining human control over critical decisions, ensuring that AI remains a tool that extends human agency rather than undermining it. Paraphrasing the chess grandmaster Garry Kasp\`arov in \citep{kasparov2010chess}, ``\textit{Weak human plus machine plus better process is superior to a strong human plus machine plus inferior process}'', he underlines the importance of efficient collaboration between humans and machines, and highlights that it is not just the strength of the individual components (human or machine) that matters, but the quality of the process and the interaction between the two. In essence, the combination of human skills, AI, and a structured approach, can outperform even the most advanced systems if they lack good design or coordination.

\vspace{\baselineskip}

The key difference between these two perspectives lies in their preserved degree of human autonomy. The delegation approach maximizes efficiency at the potential cost of autonomy, as it seeks to replace human roles with AI. In contrast, the collaborative approach values augmentation over replacement, seeking to keep humans in the loop, especially in areas where ethical considerations, emotional intelligence, and social nuances are paramount. Delegating a majority of processes to AI could lead to a society where human roles are marginalized (e.g., the irrelevance of the human being \citep{harari201821}), and individuals may feel alienated or disempowered. This could result in a sense of dehumanization, where machines make decisions that directly impact human lives without the nuanced understanding that humans bring to social contexts. 

Alternatively, using AI as a collaborative partner has the potential to empower individuals, offering new opportunities for personal and professional growth, and enabling humans to focus on aspects of life that require empathy, creativity, strategic thinking, and even social justice \citep{buccella2023ai}. The sustainability of a society that relies heavily on AI for decision-making may be called into question if such systems are not adaptable to changes in societal values and norms. A collaborative approach, however, may offer more adaptability, as it encourages continuous interaction between humans and machines, ensuring that AI systems evolve alongside human needs and priorities.

It is tempting to assume that the solution lies in achieving a balance between these two approaches. Unfortunately, such a balance seems to be difficult to achieve, in view of where it is evolving to, and how tempting it is to delegate tasks to something we think will always solve it right. Nevertheless, we envision that the role of the AI and the achievement of \texttt{TAI} could imply these points:

\paragraph{$\bullet$ Trust and Acceptance} The success of \texttt{TAI} will largely depend on the trust people place in the technology. If \texttt{TAI} is perceived as reliable and safe, it is more likely to be widely adopted. However, where it is perceived with skepticism or apprehensiveness (for example, fear that \texttt{TAI} could replace jobs on a massive scale or that it could be used for malicious purposes), resistance to its use may slow its progress. Research such as the Stanford AI Index \citep{aiindex2024} highlights how the public has concerns about bias and transparency in AI algorithms.
    
\paragraph{$\bullet$ Regulation and Governance} How governments and institutions perceive AI also influences the creation of regulatory frameworks. Doubtful views about the ethics and safety of \texttt{TAI} can lead to restrictive policies, while a positive and balanced perception can encourage regulation that promotes innovation but keeps risks under control. The development of misaligned or uncontrolled \texttt{TAI} could have catastrophic consequences, underlining the need for society to regulate technologies appropriately.
    
\paragraph{$\bullet$ Impact on Innovation} Public perception also influences investment in \texttt{TAI}, both from the private and public sectors. If AI is perceived as a tool that can improve productivity and quality of life, governments and investors are more inclined to support funding for its research and development. 
    
\paragraph{$\bullet$ Ethics and Social Welfare} The debate over the ethical benefits of \texttt{TAI}, such as its use to improve education, access to healthcare, and the reduction of inequalities, also depends on how it is perceived. The idea that \texttt{TAI} can be a positive force, rather than simply a threat, is key to achieving widespread adoption and societal benefit. The importance of developing AI technologies that are aligned with human values is crucial, reinforcing the idea that positive and ethical perceptions of \texttt{TAI} are fundamental to its success. Because of the transformative power of AI, and because it creates new possibilities and establishes new standards of well-being within society, ensuring equitable access to AI for all members of society is essential \citep{buccella2023ai}.

\vspace{\baselineskip}

Even if we strike the perfect balance between delegation and collaboration, and give AI the right role, we would still face a number of challenging human and technical-theoretical challenges, as discussed in the following points.
\section{Human Potholes in the Road to TAI}\label{human_limits}

The human factor poses significant challenges in the development of \texttt{TAI}, particularly in areas such as biases, oversight, ethical considerations, and socio-technical integration. While human involvement is critical in guiding AI research and ensuring alignment with societal goals, it can also create hurdles that may impede the development of advanced, reliable, and ethically sound AI systems.

\subsection{Cognitive and Data Biases}

One of the primary ways in which human involvement can obstruct AI development is through the introduction of cognitive biases and prejudices, which can inadvertently influence the training data used to build AI systems. Machine learning models, especially those learning from large datasets, are highly sensitive to the biases inherent in the data they are trained on \citep{mehrabi2021survey}. If the data reflects human biases, such as gender, racial, or cultural biases, AI models will likely replicate and perpetuate these biases in their outputs. The issue of encoding biases in data and overlooking critical disparities in representation poses a significant barrier to developing truly \texttt{TAI} that can operate equitably across diverse populations.

\subsection{The Information Overload}

In the contemporary era, where individuals are bombarded with vast amounts of irrelevant or misleading information, the ability to achieve clarity has become a crucial form of intellectual and practical power. As suggested in \citep{harari201821}, the challenge is no longer the access to information, but discerning what is significant and trustworthy amidst the noise. Clarity allows individuals to filter through data efficiently, make well-informed decisions, and avoid cognitive overload. This is particularly critical in fields such as politics, economics, and science, where strategic thinking and the ability to act on relevant information are paramount. Consequently, cultivating clarity is not merely a skill but a necessity for navigating the complexities of modern society. This power of clarity emphasizes the importance of critical thinking and digital literacy, enabling individuals to prioritize meaningful content and resist manipulation in a world increasingly driven by misinformation and superficiality.

\subsection{Misalignment and The Enforcement of Human Oversight}

As AI systems grow more capable, so do risks from misalignment \citep{ji2023ai,shen2023large,gabriel2024ethics,klingefjord2024human}. AI alignment aims to make AI systems behave in line with human intentions and values, and a key challenge is ensuring that AI systems consistently and robustly avoid harmful or dangerous behaviors, particularly in the case of those close to \texttt{BCAI} (because they already possess some of its capabilities e.g., such as autonomy) or deployed in safety-critical environments \citep{dalrymple2024towards}.

Therefore, human oversight in AI development is essential to ensure that systems align with ethical standards, but it can also pose challenges, especially when values between different stakeholders are misaligned. Researchers, developers, policymakers, and end-users often hold varying, sometimes conflicting, ethical perspectives on how AI should be developed and deployed in such systems (political and governance, cultural, social, communication, and information, among others). The human element in decision-making, therefore, can create a landscape where \texttt{TAI} is hindered by disagreements over ethical frameworks and acceptable trade-offs. Here we again draw attention back to the importance of reconciling different perspectives on AI Ethics (Section \ref{ssec:reconciling}). As an illustrative example, social media companies assert that they do not seek to censor the opinions of any individual on the grounds of freedom of speech. This position is understandable; however, it is important to recognize that bots, which are not human beings, do not possess the same rights as humans. Social media companies that allow non-human entities, or bots, to participate in online discourse may be held liable for the dissemination of misinformation and conspiracy theories. In this way, the algorithms that govern social networks should be held to the same standards as a newspaper editor, who is responsible for the content published under their purview.

\subsection{Risk Aversion and Inertia}

The human factor also introduces an element of risk aversion, which can slow the pace of innovation in AI. While cautious oversight is necessary to prevent the development of harmful technologies, excessive risk aversion may slow progress and delay the realization of \texttt{TAI} potential. Human decision-makers, whether in regulatory agencies, corporate boardrooms, or research institutions, are often hesitant to take bold risks, particularly in areas where the societal impact of AI is not yet fully understood. This conservatism can lead to inertia in AI development, as organizations may avoid exploring more transformative or radical AI approaches due to fear of unforeseen consequences, public backlash, or legal liabilities. AI may face hurdles due to stringent regulatory approvals, or due to the associated costs of having to comply with overly stringent regulations. Here two relevant dangers are present in our society, which can lead us to a negative perception of AI: the \textit{attention economy} and the vision of AI as a competitor:

\paragraph{$\bullet$ The Attention as a Bargaining Chip} It refers to a business model in which attention is the primary currency, monetizing users' attention \citep{franck2019economy}. Digital platforms like social media, search engines, and streaming services compete for users’ finite attention, which is then sold to advertisers as a transactional asset. The longer users stay engaged (``dopamine loop'' \citep{stjernfelt2020attention}) the more data they generate, allowing for hyper-targeted advertising campaigns and potentially greater profits. AI plays a critical role in this model by optimizing algorithms to keep users on platforms for extended periods. Through techniques such as personalized recommendations, behavioral nudging, and predictive analytics, AI ensures that content is tailored to individual preferences, often at the expense of users' autonomy and well-being. Rather than fostering intellectual growth, informed decision-making, or genuine social engagement, the AI-driven attention economy is increasingly seen as a force that undermines societal well-being by exploiting cognitive vulnerabilities, distorting information ecosystems, and fostering addictive behaviors \citep{alter2017irresistible}. 
    
We find that AI in the \textit{attention economy} not only manipulates individual attention but also distorts broader information ecosystems. Algorithms prioritize content that is most likely to engage users, often favoring sensationalism, controversy, and emotional resonance over factual accuracy or intellectual rigor. This creates ``filter bubbles'' and ``echo chambers'' \citep{pariser2011filter,terren2021echo}, where users are exposed predominantly to information that aligns with their existing beliefs and preferences. As a result, these algorithms amplify misinformation and polarize content, contributing to the fragmentation of public discourse. 
    
Another detrimental effect of AI in the \textit{attention economy} is its impact on social cohesion. Digital platforms, through their algorithms, foster an environment where users are increasingly siloed into homogeneous online communities, which are often defined not by shared interests or constructive dialogue, but by extremism and conflict. The business model incentivizes platforms to promote engagement through negative emotions, such as anger and fear, as these are more likely to drive interaction than positive or neutral content \citep{gonzalez2023social}. 
    
The mental health crisis exacerbated by the \textit{attention economy} is another significant concern. The constant torrent of notifications, content recommendations, and social comparisons driven by AI-powered platforms has been linked to increased rates of anxiety, depression, and loneliness, particularly among younger populations. Social media, in particular, has been shown to foster feelings of inadequacy and low self-esteem by promoting unrealistic standards of success, beauty, and happiness \citep{karim2020social}.

\paragraph{$\bullet$ Human vs AI} It is inevitable that human and machine capabilities are compared all the time. What makes not much sense is that this comparison is done from the perspective of human substitution by machines, instead of measuring the improvement in humans by extending their capabilities with AI. So far, one of the characteristics where the greatest distance between humans and machines is found is in reasoning and abstraction capacity. Abstract reasoning, defined as the ability to utilize known information to solve unfamiliar and novel problems, is a fundamental aspect of human cognition, observable even in toddlers \citep{piaget1952origins}. Although recent LLMs have demonstrated remarkable performance in various tasks, the extent of their true abstract reasoning capabilities remains a subject of ongoing debate \citep{bender2021dangers}. Achieving truly general AI will necessitate the development of advanced abstract reasoning capabilities, making it crucial to monitor progress in this domain \citep{aiindex2024}.

\subsection{Overreliance on Human Expertise}

Another significant issue is the potential overreliance on human expertise during AI training and deployment. While human-in-the-loop approaches \citep{valtonen2022human,mosqueira2023human} are often necessary to ensure accuracy and relevance in AI outputs, they can also limit the scalability and transformative capacity of AI systems \citep{kumar2024applications}. For example, in natural language processing or image recognition tasks, human annotators are often required to label data. However, this human involvement can introduce ambiguities, inconsistencies, errors, and subjective judgments into the training process, limiting the system’s ability to generalize beyond specific contexts or perform at scale. Moreover, human expertise itself is fallible. Experts in various fields may hold narrow perspectives, outdated knowledge, or personal biases that inadvertently shape the design and functionality of AI systems. This can lead to AI models that reflect the limitations of human knowledge, rather than transcend them to offer innovative solutions or new insights.

\subsection{The Perception of Futility of AI Ethics}

As suggested in \citep{munn2023uselessness}, the achievement of \texttt{TAI} could be also in danger when applying unhelpful ethical principles. This and other authors critique the current state of AI Ethics, arguing that ethical guidelines and principles have become ineffective and largely symbolic, highlighting several key issues:

\paragraph{$\bullet$ Proliferation of Guidelines with Limited Impact} Over the past few years, a surge of AI ethical guidelines has emerged from both the public and private sectors. However, these guidelines are often abstract, inconsistent, and lack practical enforcement mechanisms \citep{morley2023operationalising,dotan2021proliferation,kijewski2024rise}. They establish normative ideals but fail to provide actionable mechanisms or consequences for non-compliance. This means that they are frequently overwritten by economic incentives or serve as mere marketing tools, rather than genuine commitments to ethical behavior.
    
\paragraph{$\bullet$ Absence of Regulation} In \citep{munn2023uselessness} the author argues that AI Ethics is often used as a substitute for regulation. Ethical frameworks are tasked with roles they were never designed to fulfill, such as enforcing compliance and setting legal standards. Since these principles are not self-enforcing, companies can easily ignore them without facing tangible penalties. For example, technological companies have established advisory councils to oversee their AI projects. However, these bodies often lack the authority to enforce decisions internally or pause problematic projects.
    
\paragraph{$\bullet$ Cultural and Structural Issues in the Technological Industry} The culture of the technological industry, which is driven by prioritizing rapid innovation and economic growth, often sidelines ethics. The author of \citep{munn2023uselessness} suggests that many companies use ethics as a way to delay or avoid regulatory action while continuing to prioritize profit. This lack of commitment is compounded by the inadequate integration of ethics into the education of future engineers and AI developers, further perpetuating an ``a-ethical'' environment where ethical considerations are not even acknowledged during the development process.
    
\paragraph{$\bullet$ Systemic Ineffectiveness and Toothless Principles} The author of \citep{munn2023uselessness} and other references cited therein argue that without concrete enforcement mechanisms, AI Ethics will remain ``toothless'', i.e. lack enforcement power, practical applicability, or consequences for non-compliance. They emphasize that the ethical frameworks in place are insufficient for addressing the deeper social inequalities and structural issues that shape AI development. As a result, these guidelines fail to effect real change and often only serve as superficial measures to appear socially responsible.

\vspace{\baselineskip}

Noteworthy efforts have been made to imbue AI with ethics \citep{liu2022artificial}, giving rise to the field of \emph{Machine Ethics}, which tackles questions such as: Is ethics a computable function? \citep{genova2023machine}, is it possible to construct virtuous robots? \citep{gibert2023case}, or can machines be moral? \citep{sparrow2021machines}. But these efforts are made under an intense debate, since it seems difficult to dissociate oneself from the well-known maxim ``AI is not ethical, humans are''. This reflects a widely discussed concept: AI itself cannot possess inherent moral agency, so ethical considerations in AI must come from the humans who design, deploy, and regulate it. This idea has been echoed repeatedly in recent literature. A notable contribution to this sentiment can be found in \citep{russell2019human}, which argues that the morality of AI systems is fundamentally tied to the human intentions and values embedded in their design. Similarly, the author of \citep{o2017weapons} emphasizes that the ethical failures of AI systems often stem from the biases or lack of foresight in human creators.

\subsection{Socio-Technical Disparities and Digital Divide}

The human factor is also evident in the socio-technical disparities that emerge during the deployment of AI technologies \citep{kudina2024sociotechnical}. \texttt{TAI} has the potential to greatly enhance productivity, decision-making, and societal well-being. However, these benefits may not be equitably distributed due to the existence of socio-technical divides. Human decisions regarding AI accessibility, infrastructure development, and education often favor privileged regions or demographics, further widening the digital divide. For instance, AI systems developed primarily for wealthy, technologically advanced societies may not adequately address the needs of underdeveloped regions, leading to uneven adoption and missed opportunities for AI to be truly transformative on a global scale. Additionally, the human decision-making processes behind AI investment, funding, and resource allocation often prioritize short-term gains for certain sectors, while neglecting broader social concerns such as inequality or environmental sustainability.

\subsection{The Loss of Trust and Resistance}

The idea of \texttt{TAI}, especially of AI that operates autonomously or with equal human cognitive abilities, elicits significant psychological and ethical resistance among the public, policymakers, and even AI researchers. The fear of job displacement, loss of control, or even existential risks associated with AI is often heightened by dystopian narratives in media and public discourse. This widespread apprehension can lead to resistance to AI deployment, curbing innovation and delaying the integration of AI in key sectors. Furthermore, ethical concerns about the autonomy of AI systems, especially those involved in decisions that affect human lives, such as criminal justice, healthcare, or military operations, generate resistance to AI’s potential transformation of society. People often struggle to trust AI systems, especially when they do not fully understand how these systems arrive at their produced decisions. This lack of trust in AI’s decision-making capabilities, combined with the potential for errors or bias, may reduce public acceptance and hinder the adoption of \texttt{TAI}.

The misuse of generative AI deserves a special mention in this section. By this we refer to the improper or unethical application of generative AI models \citep{hagendorff2024mapping} that are designed to create content, such as text, images, audio, or video \citep{hagendorff2024mapping}. These models, while highly innovative and useful in various fields, can be exploited in harmful ways, leading to negative consequences across multiple sectors of society. One prominent concern is the creation of deepfakes, which are highly realistic yet fake media representations of individuals. These can be used to spread misinformation, damage reputations, or manipulate public opinion \citep{passos2024review}. Generative AI can also facilitate the production of fake news articles and misleading reports, contributing to the erosion of trust in information and media. Another area of misuse involves the generation of inappropriate or harmful content, such as biased, offensive, or violent materials. AI models trained on unfiltered and/or biased datasets may inadvertently generate content that reinforces harmful stereotypes or promotes discriminatory views. Additionally, generative AI poses risks in the context of intellectual property and copyright. AI-generated art, music, and writing raise complex questions about ownership and the rights of original creators, or about the ``right to be forgotten'' \citep{rosen2011right,lobo2023right} (which grants individuals the right to request the deletion of their personal data), as models can generate works that resemble or imitate existing ones without proper attribution \citep{dathathri2024scalable}. This poses a significant challenge in the field of machine learning (machine unlearning \citep{zhang2023review}), as trained models inherently retain information derived from their training data, even after the data itself has been deleted.

In their recent article \citep{marchal2024generative}, the authors shed light on the evolving landscape of generative AI misuse. They highlight that while public discourse has often focused on the fear of sophisticated adversarial attacks, their findings indicate a more widespread occurrence of low-tech, easily accessible misuses by a diverse range of actors, frequently motivated by financial or reputational incentives. Though these misuses may not always appear overtly malicious, they can have significant implications for trust, authenticity, and the integrity of information ecosystems. Additionally, generative AI exacerbates existing threats by lowering the barriers to entry and enhancing the effectiveness and accessibility of previously resource-intensive tactics. 

These insights highlight the necessity of a comprehensive approach to mitigating generative AI misuse, requiring collaboration among policymakers, researchers, industry leaders, and civil society. Addressing this challenge involves not only technical innovations but also a deeper understanding of the social and psychological factors that contribute to the misuse of these powerful technologies, expanding technological countermeasures to generative AI misuses with education and training from the earliest stages of life. 

\subsection{Diluted AI}

The advancement of AI has been so far characterized by a notable lack of a collective goal and thus global coordination, which poses significant challenges for the development of truly \texttt{TAI}. One of the primary issues is the absence of a centralized, international organization dedicated to overseeing and harmonizing AI research, development, and regulation across different nations. This fragmentation results in the duplication of efforts and resources, as various countries and institutions pursue similar AI-related objectives without collaboration or alignment. For instance, multiple countries are investing heavily in AI for defense, healthcare, and economic management, yet their research agendas are often disconnected, leading to inefficiencies and redundancies in innovation. A crucial consequence of this lack of coordination is the allocation of resources to areas that may not represent global priorities for civilization. AI has the potential to address critical global challenges, such as climate change \citep{ripple20242024}, poverty reduction, and public health. However, since national interests often drive research agendas, substantial investments are channeled into projects that serve localized or commercial objectives, rather than addressing global issues. This misalignment of priorities hinders the achievement of \texttt{TAI}, which requires a collective effort to harness the full potential of AI for the benefit of humanity as a whole.

Regrettably, the absence of a unified regulatory framework exacerbates these challenges. While several countries, notably those within the European Union and some states of the United States like California, have begun to implement AI regulations (e.g., the \textit{EU AI Act}\textsuperscript{\ref{eu_act}} and the \textit{AB 3030}\footnote{\url{https://www.gov.ca.gov/2024/09/29/governor-newsom-announces-new-initiatives-to-advance-safe-and-responsible-ai-protect-californians/?utm_source=chatgpt.com} [Accessed on December 10th, 2024].} respectively), these regulations are regional in scope and often reflect specific political, economic, or cultural concerns, despite the unifying and coordinating efforts made by the United Nations\footnote{\url{https://www.un.org/techenvoy/ai-advisory-body} [Accessed on December 10th, 2024].}, the OECD\footnote{\label{oecd_ai}\url{https://oecd.ai/en/} [Accessed on December 10th, 2024].}, or UNESCO\footnote{\label{unesco_ai}\url{https://www.unesco.org/en/artificial-intelligence} [Accessed on December 10th, 2024].}. There is no global (planetary) consensus on critical issues such as data privacy, algorithmic transparency, or ethical AI deployment. This regulatory fragmentation can lead to conflicts between nations, as divergent standards for AI safety, ethics, and governance emerge, making it difficult to create interoperable AI systems or enforce shared guidelines.

In many cases, there are also conflicts of interest at play. Nations and corporations may prioritize AI advancements that give them competitive advantages in global markets or in military applications, rather than fostering the development of AI systems that align with broader humanitarian goals. This competition can lead to ``race-to-the-bottom'' dynamics, where ethical considerations are sacrificed in the rush to achieve technological superiority. Moreover, the lack of coordinated governance mechanisms increases the risk of unequal distribution of AI benefits and widens global inequalities, as wealthier nations dominate the AI landscape while developing countries struggle to access the necessary technologies and expertise, and as demographic advantages creates an imbalance in accessing to talent or having global scientific contributions.

To overcome these barriers, the development of \texttt{TAI} requires a concerted global effort to align AI research and regulation with long-term societal goals \citep{coeckelbergh2024case}. International cooperation, perhaps facilitated by a new multilateral organization similar to the United Nations or the World Health Organization, could ensure that AI advances address pressing global challenges \citep{ripple20242024}, avoid duplication of efforts, and promote equitable access to AI technologies. Without such coordination, the promise of \texttt{TAI} may remain unfulfilled, as fragmented efforts, conflicting interests, and misallocated resources continue to hinder its development.

\subsection{Overregulation}

Despite having highlighted the need for regulation and unified guidelines at a global level, falling into excess can lead to another problem \citep{henry2022regulation}. Governments and regulatory bodies around the world are increasingly focused on establishing legal and ethical frameworks to manage the AI deployment and its applications. Although regulation is necessary to mitigate risks and ensure that AI technologies are used ethically and safely, there is a growing concern that overregulation, that is, the imposition of excessive or overly restrictive rules, could significantly hinder progress toward achieving \texttt{TAI}. Overregulation can suppress innovation, slow down technological advancements, and create bureaucratic obstacles that limit the development and deployment of advanced AI systems, as detailed below.

\paragraph{$\bullet$ Slowing Down Innovation and Research} One of the primary consequences of overregulation is the reduction in innovation. Strict regulatory frameworks can require lengthy approval processes, compliance checks, and additional documentation, all of which can delay research and development efforts. Innovators and developers may find themselves spending more time and resources on ensuring compliance than on advancing their technologies. For example, if regulations mandate that every AI system undergoes extensive ethical and safety evaluations before deployment, the time-to-market for new AI models could be significantly increased, thereby hindering the speed at which transformative advancements are realized. Moreover, the fear of regulatory penalties or noncompliance can discourage researchers and companies from pursuing ambitious AI projects, especially in areas that are perceived as high-risk, such as autonomous systems or advanced robotics. When the regulatory environment is overly stringent, it often leads to a risk-averse culture where stakeholders prefer to invest in safer, less innovative projects to avoid potential legal and financial repercussions. This dynamic can prevent the exploration of revolutionary AI advances that are critical for the development of \texttt{TAI}.

\paragraph{$\bullet$ Fragmentation and Inconsistency Across Jurisdictions} Overregulation is also associated with fragmented regulatory landscapes, particularly when different countries or regions impose their own unique sets of rules and guidelines. This lack of harmonization creates a complex environment where companies and researchers must navigate a patchwork of regulations, often facing conflicting requirements depending on the jurisdiction in which they operate. For instance, while the \textit{EU AI Act} can emphasize strict ethical compliance and privacy standards such as in video surveillance, other regions like the United States or China may adopt different approaches focusing on innovation and market competitiveness. Such discrepancies make it challenging for AI developers to create solutions that comply with multiple regulatory regimes simultaneously, leading to inefficiencies and duplication of efforts. This fragmentation limits the scalability of AI solutions and hampers international collaboration. The development of \texttt{TAI} requires a global effort, with research teams and industries across borders working together toward aligned goals. However, when regulatory systems are incompatible, they inhibit the exchange of knowledge, technology, and data. Consequently, this global disconnection undermines the collaborative efforts necessary to achieve the full potential of \texttt{TAI}.

\paragraph{$\bullet$ Bureaucratic Barriers and Compliance Costs} An environment characterized by overregulation inevitably results in increased compliance costs and bureaucratic barriers. For startups and smaller AI firms, the financial burden of meeting regulatory requirements can be prohibitive, preventing them from entering the market or scaling their technologies. This is particularly concerning, as many groundbreaking innovations often emerge from smaller entities that lack the resources of large corporations. If overregulation continues to dominate the AI landscape, the field may become dominated by a few large firms capable of absorbing the compliance costs, reducing diversity and competition in AI development (if this has not already happened). Additionally, the bureaucratic hurdles associated with overregulation can create an uneven playing field where only the most well-funded companies or those with legal expertise can thrive. This dynamic suppresses innovation and contributes to the concentration of power within a limited number of organizations, potentially undermining the democratic and open nature of AI development. Such concentration of resources and influence is likely to have a negative impact on the development of \texttt{TAI}, as fewer perspectives and ideas are represented in the process.

\vspace{\baselineskip}

Although regulation is essential to ensure the ethical and responsible development of AI, it is critical to find a balance that supports innovation rather than impedes it. A more flexible and adaptive regulatory approach that evolves with technological advancements is necessary to accommodate the dynamic nature of AI research. For instance, instead of imposing rigid rules, regulatory bodies are starting to establish regulatory sandboxes, i.e. controlled environments where companies can test new AI technologies under supervision without the burden of full compliance. This approach allows regulators to understand technology and its implications more deeply, while also enabling developers to innovate without fear of immediate regulatory constraints. Furthermore, international collaboration on AI governance is crucial to reduce fragmentation and harmonize regulations across borders. By developing global standards and best practices for AI, countries can facilitate the development of AI solutions that are compliant worldwide, promoting scalability and collaboration. As seen before, organizations like the OECD\textsuperscript{\ref{oecd_ai}} and UNESCO\textsuperscript{\ref{unesco_ai}} have already spurred initiatives aimed at fostering international cooperation on AI Ethics and governance. Nevertheless, more efforts are needed to create a unified regulatory framework that balances innovation with ethical responsibility.

\subsection{Human Obsolescence}

We have previously addressed this aspect; however, due to its significant impact, we herein devote this section to a more detailed discussion. One of the most profound dangers posed by the advancement of AI is the potential for human beings to become irrelevant in a world where AI systems outperform humans in nearly every intellectual and creative domain \citep{ferrario2024experts,lewis2024reflective}. This possibility emerges from the very essence of AI development, where the value of human input becomes increasingly diminished as AI systems mature and become self-sufficient in generating and refining knowledge, content, and insights. Next, we bring some controversial topics to the debate.

\paragraph{$\bullet$ Diminishing Role of Human Contribution} Initially, AI systems are heavily reliant on human intervention through labeling and supervision for training. Humans provide the bulk of information that AI systems process to learn patterns, make predictions, and simulate human-like responses. However, as AI technologies advance, they can start producing synthetic data and generating models without completely relying on human input (although evidence of "autophagy" has been found already in modern generative AI \citep{yang2024autophagy,xing2024ai}). These AI systems become capable of self-learning and self-improving through unsupervised and reinforcement learning, minimizing the need for further human contribution \citep{thompson2024shocking}. But the situation could become even more worrying if we ourselves facilitate this task, generating agents that are carbon copies of ourselves, and being able to simulate our behavior with similar performance \citep{park2024generative}. Once AI reaches a stage where human knowledge is no longer essential for its development, the role of humans might shift from being producers of knowledge to mere consumers of AI-generated outputs. Humans may become passive recipients of AI-generated content, products, and services, which are customized and optimized based on data that AI systems have previously accumulated. In this scenario, the human ability to create, innovate, or contribute valuable information would become increasingly irrelevant, as AI models can generate more sophisticated and tailored solutions independently (e.g., through autonomous multiagent systems \citep{wang2024survey}). Such an environment could lead to a loss of autonomy and creativity \citep{prunkl2024human}, as the skills and talents that once defined human uniqueness are replaced by AI capabilities that exceed human limitations.

\paragraph{$\bullet$ The Ethical and Psychological Implications of Irrelevance} The notion of becoming irrelevant in a society where AI systems dictate economic and creative production is not merely a technical issue; it also poses deep ethical and psychological challenges \citep{harari201821}. If AI assumes roles that provide meaning and value to human life (such as work, creativity, and knowledge production, among others), there is a risk that individuals may experience a sense of disconnection and purposelessness. Historically, work and creation have been central to human identity and self-worth; the absence of these elements could lead to widespread existential crysis, as humans struggle to find their meaning and purpose when their contributions are felt as no longer needed. Furthermore, this hypothesis would reinforce a feedback loop: as AI systems continue to refine their outputs based on human consumption patterns, they further entrench the human role as mere consumers. Humans provide the data that allows AI to tailor increasingly sophisticated products, while their role in shaping or influencing these products diminishes. The dominance of AI-generated realities may also blur the lines between authentic human and artificial experiences, leading to an environment where distinguishing between what is real and what is algorithmically constructed becomes challenging.

\vspace{\baselineskip}

To avoid human obsolescence in the future, it is essential to promote AI development frameworks that emphasize the complementarity between AI and human abilities rather than the replacement of human roles. AI should be designed not to dominate every aspect of human life, but to augment human creativity and decision-making. Developing AI systems that are dependent on human collaboration ensures that human input remains valuable and relevant. This requires a shift in how AI Ethics and governance are structured. Instead of focusing solely on minimizing harm or managing AI risks, there must be an emphasis on promoting AI-human symbiosis. Policies should prioritize AI applications that empower humans to take on more meaningful and creative roles, ensuring that technology enhances rather than replaces human agency. Encouraging collaboration between AI developers, ethicists, policymakers, and the public is critical to creating a future where AI supports human flourishing rather than rendering it obsolete.

\subsection{The AI-tocracy}

The concepts of \textit{AI-tocracy} \citep{beraja2023ai}, ``algocracy'', algorithmic governance, or the ``dictatorship of the algorithm'', among others, refer to algorithms that increasingly influence or control political and social systems, eventually becoming more relevant than human themselves \citep{fioriglio2015freedom,danaher2016threat}. While algorithmic systems can optimize and automate decision-making processes at unprecedented scales, they also pose significant risks to democratic values and human autonomy \citep{engin2019algorithmic,coeckelbergh2023democracy}. The evolution toward an AI-tocracy threatens to undermine human agency \citep{mittelstadt2016ethics}, create opaque decision-making processes, and concentrate power in the hands of a few technocratic elites or corporations, as the following examples demonstrate.

\paragraph{$\bullet$ The Rise of AI-tocracy: Efficiency at the Cost of Transparency} In an AI-tocracy, algorithms are not merely tools that aid human governance; they become the decision-makers themselves, effectively substituting human judgment. This shift promises efficiency by processing vast amounts of data to optimize various societal functions, such as law enforcement, social services, and economic management. However, as algorithms gain authority, the mechanisms of governance become opaque. Unlike human legislators or judges, algorithms do not offer explanations or rationale for their decisions in a way that is easily understood by society. This lack of transparency not only obscures how decisions are made, but also undermines accountability. The notion of an algorithmic dictatorship emerges when the rules encoded within these systems become the ultimate authority, with humans relegated to the role of passive executors. In such a scenario, society is governed by decisions that are often based on inscrutable models, inaccessible to public scrutiny or democratic oversight. As these systems grow in complexity, they become uncontrollable black boxes, making it difficult for anyone -- experts, legislators, or the general public -- to challenge or appeal their decisions.

\paragraph{$\bullet$ The Loss of Human Autonomy and Agency} The shift toward an algorithmically governed society raises significant concerns about human autonomy. When algorithms dictate access to resources, determine legal outcomes, or prioritize healthcare needs, humans become increasingly subordinate to technological systems. This turns individuals into subjects who have little influence over the rules that govern their lives, as algorithmic models prioritize efficiency and productivity over moral or ethical considerations. The concept of algorithmic governance amplifies this risk. For instance, these algorithms often perpetuate existing biases, such as racial or socio-economic discrimination. The consequence is a loss of individual autonomy, as people are judged and categorized by opaque criteria that they cannot contest or influence. When AI systems become arbiters of human behavior, citizens lose the ability to shape their own destiny, leading to a society where human rights are demoted to the dictates of the algorithms.

\paragraph{$\bullet$ Concentration of Power and Technocratic Elites} As algorithmic governance becomes more prevalent, power inevitably concentrates on the hands of those who design, control, and own these technologies, typically large corporations and a small group of technocratic elites. This creates a significant power imbalance in society, where the public has little influence over the decisions made by algorithmic systems. In such a structure, democratic processes are bypassed, and policies are shaped by corporate interests that prioritize profit or efficiency over social welfare and ethical standards. This power scheme reinforces the concept of the ``dictatorship of the algorithm'', where algorithmic rules, often designed by private entities, become the \textit{de facto} law. For example, technological companies that develop surveillance technology and predictive policing algorithms can effectively shape public policy without any democratic mandate. The result is a society where corporate algorithms dictate critical aspects of human life (from credit scoring and employment decisions to policing and public health) without meaningful oversight or input from the people affected by these systems.

\paragraph{$\bullet$ The Erosion of Democratic Values} Algorithmic governance poses a direct threat to democratic values, such as transparency, accountability, and participation. Traditional democratic institutions, like parliaments and courts, rely on open deliberation and public involvement. In contrast, AI-driven systems often operate behind closed doors, governed by proprietary algorithms that are shielded by corporate secrecy. When crucial decisions about welfare, justice, and resource allocation are automated, citizens lose their voice in the political process, and their ability to participate in governance diminishes. Furthermore, the shift toward an AI-tocracy can exacerbate existing social inequalities. As mentioned before, algorithms tend to reproduce and amplify biases present in their training data, leading to discriminatory outcomes that are often difficult to identify or contest. This challenges the fundamental democratic principle of equality before the law, and further alienates marginalized communities from decision-making processes. In an environment where algorithmic decisions go unquestioned, it becomes difficult to ensure fairness, justice, and inclusivity \citep{segev2024artificial}.

\vspace{\baselineskip}

To prevent this dystopian scenario it is essential to implement robust ethical frameworks and regulatory oversight for AI systems. Governance models must ensure that algorithms are transparent, auditable, and accountable to the public. This can be achieved through legislation that mandates explainability in algorithmic decision-making processes, giving citizens the right to challenge and understand decisions that affect their lives. Additionally, promoting algorithmic literacy is crucial. Educating the public about how algorithms function and the potential biases they carry can empower citizens to engage critically with AI systems and advocate for more inclusive and ethical AI governance. Moreover, involving diverse stakeholders (including ethicists, sociologists, and representatives from marginalized communities) in the development and oversight of algorithmic systems can help ensure that these technologies serve the interests of all, not just the few who control them.

We have already examined the human-related challenges -- or ``potholes'' -- on the road to achieving \texttt{TAI}, focusing on issues such as ethics, governance, and societal readiness to embrace this technology. In the following section we turn our attention to the technical hurdles that must be overcome for \texttt{TAI} to unleash its full potential, exploring the critical gaps and algorithmic obstacles that arise from this road.
\section{Technical Potholes in the Road to TAI}\label{techsci_limits}

The journey toward \texttt{TAI} is fraught with numerous technical challenges that must be addressed to ensure both feasibility and safety along the way. In this section, some of these challenges are highlighted for their significant impact on this journey.

\subsection{The Capacity to Detect New Emerging Abilities}\label{emergence_new_ab}

The \textit{emergence theory} \citep{anderson1972more,bedau1997weak,holland2000emergence} suggests that in a system, as elements interact with each other dynamically, new properties and behaviors may emerge that are not present in the individual components. These arisen behaviors are, in many cases, unpredictable from the characteristics of the parts. Following this theory, it has been speculated that, if a sufficiently complex and sophisticated architecture could be achieved, artificial life could emerge \citep{gershenson2023emergence,alife2024}, even a superior AI could sprout as an emergent phenomenon from the interaction of many simpler AI components (as Minsky proposed in \citep{minsky1988society} with his formulated \textit{Society of Mind} theory, or as suggested in \cite{garbus2024emergent}). In this direction, the author of \citep{wolfram2003new} demonstrates how simple rules in cellular automata can produce highly complex patterns, also suggesting that complexity does not necessarily require complicated systems or initial conditions.

However, this remains a topic of intense debate, as emergence does not inherently ensure the development of capabilities like deep reasoning or self-awareness. While emergent properties are a common manifestation in complex systems, human consciousness, and general intelligence have not been reproduced simply by increasing the complexity of a system. Moreover, perhaps the ``secret'' lies not in the evolution of separate entities, but in a population and its ecosystem. Organisms inhabit ecosystems that consist of populations of their own species, other species, and an environment made up of abiotic elements and processes. These ecosystems offer a variety of niches (distinct ways of surviving and interacting) that can emerge, vanish, and evolve over time \cite{alife2024}.

Moreover, we may not be able to detect this emergence in the event it eventually occurs. Numerous studies have recently argued that LLMs demonstrate emergent abilities \citep{wei2022emergent,zhang2024intelligence}, meaning that they can unexpectedly and abruptly exhibit new capabilities as their scale increases. This has prompted concerns that even larger models might develop surprising and potentially uncontrollable new skills. A recent study \citep{li2024geometry} has revealed unexpected geometric structures within LLMs such as \textit{ChatGPT}, shedding light on how these models organize information, mirroring brain-like patterns once believed to be unique to biological systems. This breakthrough represents a significant step forward in unraveling the internal workings of AI, which have long remained enigmatic. The authores of \citep{schaeffer2024emergent}, however, challenges this perspective, suggesting that the perceived emergence of new capabilities often reflects the nature of the benchmarks used for evaluation, rather than being an intrinsic characteristic of the models themselves.

The \textit{emergence theory} is notably controversial (e.g. this statement\footnote{\url{https://situational-awareness.ai/} [Accessed on December 10th, 2024].}) in the field of AI due to attempts at establishing analogies with the human mind and the emergence of its capabilities. Without intending to delve deeply into the \textit{emergence theory} in AI—an endeavor that would warrant a dedicated study—we find it valuable to pause here to reflect on the significance of this theory at both the human and artificial levels. To this end, in what follows we highlight the challenges associated with detecting the spontaneous emergence of such capabilities:

\paragraph{$\bullet$ The Opacity} As AI systems increase in complexity, e.g. with the rise of deep learning models containing billions of parameters, their internal workings become highly opaque even to their developers. This ``black box'' nature makes it difficult to trace the direct cause of certain outputs or behaviors. If new capabilities were to emerge within such systems, they might manifest subtly at first, escaping detection because their origins would be hidden within layers of computations too complex to be analyzed. This makes the detection of emergent behaviors an ongoing challenge in AI interpretability and transparency.
    
\paragraph{$\bullet$ The Unintentionality} Unlike traditional software systems, where new functions are deliberately encoded, the capabilities of AI models are not explicitly programmed. Instead, they are learned from vast amounts of data. Emergent properties could arise as an unintended byproduct of this learning process, not as the result of a deliberate design. Since these capabilities would not be part of the intended functionality, there would be no obvious signals to alert developers or users to their presence, making them harder to be recognized until they reach a level of expression that directly impacts their observable behavior.
    
\paragraph{$\bullet$ the Incrementality} Emergent capabilities are likely to manifest incrementally, starting in ways that might not significantly diverge from expected behaviors. This subtle development could make it difficult to detect early stages of emergent intelligence or advanced functionalities. For instance, an AI system might begin to display a deeper understanding of context or make more sophisticated connections between disparate pieces of information, without overtly demonstrating a drastic shift in performance. Such gradual changes can easily be overlooked in experimental evaluations, especially if the testing phase focuses on predefined tasks and task performance rather than on the exploratory analysis of potential abnormal behaviors.
    
\paragraph{$\bullet$ The Evaluation} Most AI systems are evaluated based on established benchmarks that test specific capabilities or tasks. Emergent behaviors, by definition, are unanticipated and may not align with the metrics used for evaluating AI performance. This creates a blind spot in traditional evaluation frameworks, as they may not be designed to detect novel capabilities outside of their intended scope. Consequently, emergent properties might not be observed until they have fully developed and produce unexpected results in real-world applications, far beyond the narrow confines of pre-set evaluations.
    
\paragraph{$\bullet$ The Time Management} Another difficulty in detecting emergent capabilities is that their manifestation may occur after long periods of interaction or learning. AI models, particularly those used in dynamic environments such as reinforcement learning, may develop new abilities only after prolonged exposure to complex scenarios. These delayed effects mean that emergent properties might remain dormant for significant periods before revealing themselves, further complicating efforts to monitor and anticipate such changes in real-time.
    
\paragraph{$\bullet$ Expectations vs Randomness} In complex systems, behavior that deviates from expectations may initially be dismissed as noise or anomalous output, rather than as a precursor to emergent capabilities. Without a clear framework to distinguish between unintentional errors and the early stages of emergence, detecting these new capacities may be delayed until they have fully formed. This challenge is exacerbated by the stochastic nature of many AI models, where randomness is often part of their inner functioning, making it even harder to identify the significance of novel behaviors early on.

\subsection{The Data Paradox}\label{data_paradox}

AI models necessitate the utilization of copious quantities of data to accurately discern patterns, make predictions, and generalize to novel tasks (as the generalization ability of AI is limited by the representativeness of training data \citep{vapnik2015uniform}). The greater the diversity and abundance of the data, the more effectively the model can comprehend intricate relationships within it. Nevertheless, the acquisition of authentic, real-world data may be constrained by several factors, including concerns regarding distribution shifts or rare events that can drastically degrade performance \citep{sato2021survey}, privacy, financial costs, or accessibility, particularly in highly specialized domains such as healthcare or autonomous driving. To surmount these constraints, the generation of synthetic data seems to be imperative. It enables the fabrication of extensive, heterogeneous datasets that can emulate real-world scenarios without the constraints of data scarcity. By augmenting training datasets, synthetic data can enhance model robustness, facilitate training in scenarios where data collection is impractical, and address issues such as data imbalance. However, meticulous management is essential to circumvent reinforcing biases or creating overly simplified patterns. But this is not a bed of roses.

The generation of synthetic data can cause AI models to collapse due to the compounding of errors and biases that occur when models are repeatedly trained on AI-generated data rather than real-world data \citep{shumailov2024ai,xing2024ai}. Over time, this process leads to a loss of data diversity and richness, error propagation, simplification and homogenization, as synthetic data lacks the complexity and unpredictability found in natural environments. Consequently, models become increasingly overfitted to simplified patterns, and feedback loops reinforce inaccuracies, reducing the model's ability to generalize to new or real-world scenarios. This degradation in performance is referred to as \emph{model collapse}, as the AI loses its robustness and reliability. Finally, the authors of \citep{thompson2024shocking} highlight the significant presence of machine-generated translations across various languages and domains, raising important questions about the reliability, quality, and ethical implications of such content. This has significant implications for AI systems that are predominantly trained on web-scraped data: introducing inconsistencies, translation errors, and cultural inaccuracies into AI training datasets, and potentially impacting the performance of models in tasks such as language translation, summarization, and sentiment analysis. Moreover, the pervasive use of machine-translated data could inadvertently lead to a feedback loop, where models are trained on machine-generated content, perpetuating existing biases or errors. This highlights the importance of curating high-quality and diverse training data to ensure the robustness and reliability of AI systems.

\subsection{World Modeling}\label{world_mod}

The \textit{World Modeling} approach \citep{xiang2024language} is attracting much attention, promising to be the next step in modeling and AI. World models in AI aim to construct internal representations of an environment, enabling an AI-based system to understand, predict, and interact with the world in its surroundings. Hence, world models support decision-making, planning, and simulating potential future scenarios. By learning the underlying dynamics of an environment, world models allow AI systems to perform tasks more efficiently, even in the absence of direct data stimuli or explicit supervisory signals.

The minimization of the cross-entropy is critical component in training world models. By reducing cross-entropy, models can improve their ability to predict state transitions and make better decisions in simulated or real environments. These capabilities are valuable in fields such as robotics, autonomous vehicles, and reinforcement learning, where understanding and anticipating environmental dynamics are essential. However, while cross-entropy is useful in fine-tuning predictions and simulating environments, achieving \texttt{BCAI} capable of reasoning like a human or adapting to unseen contexts still requires advances beyond simply optimizing a loss function based on Information Theory. \textit{World Modeling} is a step in that direction, but there are still open challenges in knowledge representation, causal reasoning, and abstract understanding of the world. Real-world environments are incredibly complex, and simply minimizing cross-entropy does not always lead to a deep understanding of the underlying dynamics of the environment. Models may learn superficial patterns that work well in the short term, but fail when faced with new or unseen situations. In complex or nondeterministic scenarios, cross-entropy can be difficult to minimize effectively due to high uncertainty and noise in the data. Models need to be able to handle not only predictions based on historical data, but also the generation of abstract representations of unknown dynamics.

\subsection{Sustainability, Physical Constraints, and Alternatives}\label{hardware}

As AI models become more sophisticated and larger in scale, the energy consumption required for training and deploying these systems increases exponentially. This gives rise to concerns regarding the sustainability of AI development \citep{hacker2024sustainable}, given that the environmental impact and economic costs associated with maintaining such infrastructure are becoming considerable. 

Under these circumstances a a dormant technology is coming to a new life: the nuclear power \citep{cha2024potential}. Some global corporations are interested in developing and constructing Small Modular Reactors (SMRs), which reflects the accelerating revival of nuclear energy as a critical component of future energy strategies. Nuclear power, once seen as a controversial energy source due to safety concerns and waste management issues, is now being reconsidered as a reliable, carbon-free alternative to fossil fuels. SMR technology, which offers scalability, enhanced safety features, and reduced construction times, plays a pivotal role in this resurgence. 

However, challenges remain in terms of regulatory approval, public acceptance, and long-term waste management. As the nuclear industry moves forward, it must address these concerns to ensure the sustainable and responsible development of SMRs. In this context, this represents a key step in legitimizing nuclear energy's role in the global energy landscape, potentially encouraging other private companies to follow suit. The resurgence of nuclear energy, led by SMRs, could prove to be a game-changer in achieving carbon neutrality and energy security. Despite this, there is still hope for safer alternatives, as shown in recent studies on geothermal projects utilizing supercritical water \citep{meyer2024permeability}.

Beyond energy consumption, there are physical constraints on the miniaturization and optimization of microchips, known as \texttt{Moore's Law}\citep{moore1965moore}. This predicts that the number of transistors on a chip will double approximately every two years, resulting in more powerful processors. However, this trend is beginning to slow down as the industry approaches the atomic scale, where further miniaturization faces quantum limitations. Besides, the reliance on rare and specialized materials, such as silicon and rare earth metals, for hardware production creates vulnerabilities in the supply chain. Geopolitical factors, resource scarcity, and economic volatility further exacerbate these challenges, posing potential bottlenecks for \texttt{TAI}.

\subsection{A Stronghold for a Few}\label{stronghold}

The field of AI is increasingly becoming a small and closed domain, in which only a few companies and individuals can engage in cutting-edge research \citep{owid-ai-impact}. This growing exclusivity is driven by several factors, including the increasing specialization and complexity of knowledge, the restricted access to large-scale datasets, the different access to talent, and the escalating costs and requirements of computational resources. As these barriers rise, the AI landscape is progressively consolidating, leaving only a select few entities with the capacity to advance the field, and appropriating AI in not always direct and transparent ways \citep{ahmed2023growing,verdegem2024dismantling}:

\paragraph{$\bullet$ Specialization and Complexity of Knowledge} AI research has evolved into a highly specialized field. The theoretical foundations, algorithms, and technical skills required to innovate at the forefront of AI are becoming more complex, making it increasingly difficult for individuals or smaller organizations to keep pace. The rapid advancements in AI require not only deep expertise in Computer Science, but also interdisciplinary knowledge, including Neuroscience, Linguistics, and domain-specific applications. Also in Mathematics \citep{bengio2024machine}. This complexity creates a steep learning curve for those entering the field, and restricts the capacity of smaller research teams or independent researchers to make meaningful contributions to AI innovation. Moreover, as AI becomes more advanced, the focus has shifted toward niche areas such as reinforcement learning, natural language understanding, and multimodal systems, which demand a highly specialized set of skills. As a result, AI research is becoming increasingly insular, accessible only to those with highly advanced training and significant resources.
    
\paragraph{$\bullet$ Restricted Access to Data} Data are the lifeblood of AI research, and the quantity and quality of data required to train state-of-the-art models has grown exponentially. However, access to the large datasets needed to drive modern AI models is increasingly restricted to a few large organizations. Many of the most valuable datasets, ranging from social media interactions to proprietary business data, are controlled by a handful of corporations. These companies show a significant strength in terms of data accumulation and usage. This disparity creates a competitive advantage for major technological companies, while universities, smaller enterprises, and independent researchers face significant challenges in acquiring similar datasets. Publicly available datasets, while helpful, often lack the scale and diversity necessary to train high-performance AI systems. As a result, those without access to proprietary or large-scale data sources are left behind, unable to compete in the race for \texttt{TAI}.
    
\paragraph{$\bullet$ Escalating Computational Requirements and Costs} AI, particularly with the advent of deep learning models, demands enormous computational power. The most advanced AI models, such as LLMs and reinforcement learning agents, require not only vast amounts of data, but also substantial processing capacity. Training of these models often involves massive clusters of parallel processing units available in computing infrastructure that is only accessible to a few wealthy corporations and well-funded academic institutions. The high computational costs associated with training cutting-edge AI models make it difficult for smaller organizations and individuals to engage in meaningful research. Additionally, the costs of cloud computing services have continued to rise, particularly for intensive AI workloads. This has led to a growing divide between those who can afford to experiment with large-scale AI models and those who cannot, further concentrating AI innovation in the hands of a few.
    
\paragraph{$\bullet$ Concentration of AI Talent and Expertise} As AI becomes more specialized and resource-intensive, the pool of talent capable of conducting high-level AI research is also narrowing. Top AI researchers are increasingly concentrated in a small number of large technological companies and elite academic institutions. These organizations offer salaries and resources that are out of reach for smaller players, creating a talent drain that further consolidates AI expertise within a small number of entities. Moreover, the specialized nature of AI research means that even within large organizations, only a limited number of experts have the knowledge and skills necessary to develop cutting-edge models. This talent concentration not only suppresses innovation outside these elite groups, but also reinforces the power and influence of the few organizations that can afford to attract and retain top-tier AI talent \citep{korinek2023market}.
    
\paragraph{$\bullet$ Intellectual Property and Proprietary Models} AI is becoming progressively more competitive, and the sharing of knowledge and resources that characterized the early days of the field is diminishing. Many of the most advanced AI models are proprietary, and companies are increasingly reluctant to share their algorithms, datasets, or research findings. While open-source initiatives have democratized many aspects of AI development, the most impactful models and tools are often kept behind corporate walls. This trend toward closed-source AI further limits access for independent researchers and smaller organizations, as they are unable to replicate or build upon the latest advancements. Intellectual property laws also play a role in creating barriers to entry, as patents on AI technologies can prevent competitors from exploring similar innovations. This reinforces the dominance of a few large companies that hold the rights to key AI technologies, further concentrating control over the direction and development of the field.

\subsection{The Duality of Theoretical Foundations of Computation and AI}\label{dual_nat}

\begin{quote}
    ``\textit{The question of whether machines can think is about as relevant as the question of whether submarines can swim.}''
    \\
    --- \textit{Edsger W. Dijkstra}
\end{quote} 

The Dijkstra's quote underscores the notion that comparing the ability of machines to think with human's cognitive processes is as misplaced as comparing the ability of submarines to swim with that of a fish. He then challenged the relevance of the question by highlighting a fundamental misalignment in analogies. We would like to reinterpret this quote to critique the way that many individuals approach the concept of AI. 

Dijkstra argued that asking whether machines can ``think'' is a fundamentally misguided question, as it applies a biological concept (human thought) to machines, in much the same way that it would be inappropriate to ask whether submarines ``swim'', a term that describes a biological function. Submarines move underwater, but in a manner entirely distinct from how living organisms swim. By analogy, machines can perform tasks that resemble human thought processes, yet they do so using methods that differ fundamentally from human cognition. Dijkstra’s point emphasizes that the discourse on AI should not be concerned with whether AI can think in the human sense, but rather with whether they can autonomously and effectively perform complex tasks. His statement highlights the importance of recognizing that the value and utility of machines do not depend on their capability to replicate human thinking and skills identically, but on their ability to achieve similar outcomes through different means. In light of recent AI advances, we have seen that it is not essential to possess an AI with exceptionally sophisticated capabilities to continue driving progress. In this sense, while the human brain has undeniably inspired much of the development of contemporary AI (e.g. neural networks) \citep{ren2024brain,wang2024brain}, it does not necessarily need to serve as the definitive blueprint for its design. The brain can provide a reference for what intelligence achieves (e.g. reasoning, learning, and decision-making), but not necessarily how these capabilities must be implemented. This distinction echoes Dijkstra's quote where this reinterpretation is intended to show how AI systems can diverge from biological intelligence in structure and function, while achieving equally transformative results. By freeing AI development from strict adherence to neural emulation, researchers could explore novel architectures and approaches that optimize for different goals, constraints, and capabilities, potentially surpassing the limitations of human cognition in certain contexts. This conceptual flexibility underscores the potential of AI as an innovation not constrained by biology, but rather informed by it \citep{zhao2022inspired,gershman2024have}.

Therefore, AI could transition to \texttt{TAI} without the necessity of progressing through \texttt{BCAI} (as illustrated by Dijkstra's quote). However, it is indeed the case that the latter pathway is the one most commonly associated with the concept of \texttt{TAI}. Consequently, this part will examine some of the most widely discussed theoretical arguments both in favor of and against this possibility.

The development of \texttt{BCAI} may be underpinned by a complex interplay of theoretical foundations and inherent limitations derived from fundamental theorems in Computer Science, Mathematics, and Information Theory. These frameworks provide essential insights into the capabilities and constraints of computational systems, shaping the possibilities of achieving \texttt{BCAI}. Therefore, this part explores their dual nature, examining how they support or challenge the realization of \texttt{BCAI}. By analyzing their implications, this discussion aims to provide a balanced perspective on the theoretical landscape of \texttt{TAI}, emphasizing areas where innovation and practical trade-offs are essential for progress. The synthesis of these diverse perspectives (see Table \ref{tab:theorems_bcai}) reveals not only the immense potential of \texttt{TAI} but also the profound challenges it faces as an inherently computational endeavor.

\begin{table}[H]
\centering
\renewcommand{\arraystretch}{1.5}
\resizebox{0.7\textwidth}{!}{ % Scale to 80% of text width
\begin{tabularx}{\textwidth}{@{}>{\centering\arraybackslash}p{3cm}X>{\centering\arraybackslash}p{3cm}@{}}
\toprule
\textbf{Theorem} & \textbf{Key Contribution and Limitation for \texttt{BCAI}} & \textbf{Role} \\ 
\midrule
\texttt{Turing Completeness} & Demonstrates that any computable process can be simulated by a Turing machine, enabling universal computation. However, it cannot solve non-computable problems, such as the Halting Problem, which imposes inherent limits on predictability. & Foundational enabler of computation and \texttt{BCAI} feasibility. \\ 
\texttt{Universal Approximation Theorem} & Establishes that neural networks can approximate any continuous function, supporting \texttt{BCAI} in modeling complex cognitive tasks. Practical issues like overfitting and training inefficiencies, however, limit scalability. & Supports flexibility and adaptability in learning systems. \\ 
\texttt{Bayesian Inference} & Provides a framework for reasoning under uncertainty, essential for \texttt{BCAI} adaptability. Yet, the computational cost in high-dimensional spaces can hinder its real-world application. & Enables probabilistic reasoning and robust decision-making. \\ 
\texttt{Hutter's Universal AI} & Defines a theoretically optimal agent for decision-making in computable environments. However, its reliance on computationally intractable methods, like Solomonoff induction, makes implementation infeasible. & Provides an idealized framework for \texttt{BCAI} optimization. \\ 
\texttt{Algorithmic Information Theory} & Offers principles for efficient data representation and generalization, essential for learning. Nonetheless, undecidability in finding minimal representations constrains \texttt{BCAI}'s optimization. & Enables efficient encoding and generalization but highlights inherent limitations. \\ 
\texttt{G\"odel's Incompleteness Theorem} & Highlights that some truths in formal systems are undecidable, limiting \texttt{BCAI}'s ability to achieve complete reasoning. Encourages modular and approximate reasoning approaches. & Limits comprehensive reasoning and universal consistency. \\ 
\texttt{Halting Problem} & Demonstrates that no algorithm can universally predict program termination, impacting \texttt{BCAI}'s reliability in dynamic systems. Promotes modular and bounded problem-solving to mitigate this limitation. & Highlights unpredictability and promotes bounded approaches. \\ 
\texttt{Complexity Theory} & Suggests that many real-world problems are computationally intractable if $P \neq NP$, posing scalability challenges for \texttt{BCAI}. Approximation methods provide a workaround for bounded domains. & Exposes scalability challenges and promotes heuristic solutions. \\ 
\texttt{Shannon’s Information Theory} & Enables efficient data compression and communication, vital for large-scale AI systems. However, limits on channel capacity and noise robustness can constrain reliability. & Supports efficient data handling but imposes communication limits. \\ 
\texttt{Rice's Theorem} & States that non-trivial properties of program behavior are undecidable, preventing \texttt{BCAI} from fully analyzing or verifying the behavior of itself or other systems. Encourages heuristic and probabilistic approaches. & Constrains self-analysis and system verification. \\ 
\bottomrule
\end{tabularx}
}
\caption{Comparative overview of the most accepted theoretical foundations of Computation and AI, with their roles and their impact on achieving \texttt{BCAI}.}
\label{tab:theorems_bcai}
\end{table}

Next, we go into more detail on each of these theorems/theories:

\paragraph{$\bullet$ Turing Completeness} It establishes the theoretical foundation for universal computation \citep{turing1936computable}. A Turing-complete system can simulate any computation given sufficient time and resources. This universality implies that any algorithmic cognitive process, including those underlying human intelligence, can, in principle, be implemented on a computational system. Turing completeness thus supports \texttt{BCAI} by asserting that all computable aspects of human reasoning, learning, and problem-solving can be reproduced. However, it is worth noting that this theory applies only to computable functions, leaving undecidable problems beyond its scope. Nonetheless, Turing completeness underpins the theoretical possibility of \texttt{BCAI}, as human cognition, if reducible to algorithmic models, can be executed on Turing machines or equivalent systems \citep{turing2009computing}. While Turing completeness guarantees the theoretical feasibility of implementing any computable process, several practical challenges remain for achieving \texttt{BCAI}, such as the efficiency of computation (e.g. time and memory) or that only applies to computable functions.

\paragraph{$\bullet$ Universal Approximation Theorem} This theorem first formalized in \citep{cybenko1989approximation} and later generalized in \citep{hornik1991approximation}, states that a neural network with a sufficient number of neurons in a single hidden layer can approximate any continuous function on a compact subset of $\mathbb{R}^n$, i.e. the n-dimensional Euclidean space. This result demonstrates the theoretical capacity of neural networks to model complex relationships and dynamics inherent in cognitive processes. It supports \texttt{BCAI} by providing the mathematical foundation for deep learning models to approximate and potentially surpass human-level capabilities in various domains. However, practical challenges, such as overfitting, training inefficiencies, and data availability, limit its immediate applicability to \texttt{BCAI}. Nonetheless, it highlights the potential of neural networks to approximate complex functions, which does not directly address the ability to generalize cognitive functions akin to human intelligence.

\paragraph{$\bullet$ Bayesian Inference} Rooted in Bayes’ Theorem \citep{bayes1763essay}, it offers a framework for reasoning under uncertainty by updating probabilities as new evidence becomes available. This methodology is critical for \texttt{BCAI}, as it enables systems to adapt dynamically to incomplete or noisy data, an essential feature for real-world applications. Bayesian inference supports \texttt{BCAI} by formalizing how can process uncertain environments more systematically and effectively than humans, who are prone to cognitive biases. However, it also introduces limitations, as priors must be carefully defined, and the computational cost of Bayesian methods can be prohibitive in high-dimensional scenarios. Despite these challenges, Bayesian reasoning is fundamental for creating adaptive and robust \texttt{BCAI} \citep{pearl2009causality,gerstenberg2024counterfactual}. 

\paragraph{$\bullet$ Algorithmic Information Theory} It defines the complexity of data as the length of the shortest program that can generate it \citep{solomonoff1964formal,kolmogorov1965three}. This principle supports \texttt{BCAI} by formalizing how intelligent systems can identify and generalize patterns efficiently. It aligns with Occam’s Razor \citep{blumer1987occams}, prioritizing simpler models that generalize better. However, the theory imposes a fundamental limitation: computing the minimal program is undecidable. This constraint means that while \texttt{BCAI} could approach optimal generalization, it cannot achieve perfect model minimality. Nonetheless, algorithmic information theory provides a foundational framework for building systems that learn efficiently.

\paragraph{$\bullet$ Hutter's Universal AI} The Marcus Hutter’s Universal AI framework \citep{hutter2005universal} introduces the AIXI model as a theoretical \texttt{BCAI} agent. AIXI combines Solomonoff induction \citep{solomonoff1964formal} with reinforcement learning to achieve optimal decision-making in any computable environment. By leveraging simplicity and Bayesian principles, AIXI represents a theoretically \texttt{BCAI} capable of surpassing human intelligence in well-defined tasks. However, its reliance on Solomonoff induction makes AIXI computationally intractable, as it involves summing over all possible hypotheses weighted by complexity. Despite this, Hutter's work provides a formal definition of \texttt{BCAI}, reinforcing the theoretical possibility of \texttt{BCAI} while highlighting significant practical barriers.

\paragraph{$\bullet$ The G\"odel's Incompleteness Theorem} It demonstrates that in any consistent formal system capable of expressing basic arithmetic, there are true statements that cannot be proven within the system \citep{godel1931formal}. This theorem imposes a fundamental limitation on \texttt{BCAI} by highlighting the inherent incompleteness of formal reasoning frameworks. While it does not negate the possibility of \texttt{BCAI}, it suggests that such systems may encounter problems analogous to G\"odel's sentences (true but undecidable). This limitation implies that \texttt{BCAI} could not fully replicate or surpass human reasoning in every domain, as it may fail to address inherently undecidable questions. This theorem challenges the idea that the universe, human cognition, or mathematics can be completely understood through mechanistic or algorithmic means.

\paragraph{$\bullet$ Chaitin's Incompleteness Theorem} It extends G\"odel's work to algorithmic information theory, demonstrating that there are limits to what can be proven about the complexity of programs within formal systems \citep{chaitin1974information}. This imposes a constraint on \texttt{BCAI} by highlighting the impossibility of finding the shortest or most efficient representations for all problems. While \texttt{BCAI} could approximate solutions, it cannot guarantee optimality in all cases. Chaitin's Theorem underscores the trade-offs between precision, computational cost, and practical feasibility in \texttt{BCAI} development. 

\paragraph{$\bullet$ Rice's Theorem} It poses a fundamental challenge to the realization of \texttt{BCAI} by establishing that any non-trivial property of a program’s behavior is undecidable \citep{rice1953classes}. This result implies that a \texttt{BCAI} system cannot universally and algorithmically determine whether another program (or even itself) exhibits certain desirable properties, such as safety, ethical alignment, or performance optimization. For instance, a \texttt{BCAI} attempting to analyze or verify its own modifications during a process of recursive self-improvement would encounter inherent computational barriers, as verifying whether such modifications improve functionality or maintain alignment with human values is undecidable for non-trivial cases. This limitation extends to the prediction of the behavior of external agents or systems, particularly in complex, dynamic environments, where \texttt{BCAI} systems are expected to generalize and adapt. Consequently, Rice's Theorem highlights the need for \texttt{BCAI} to rely on heuristic methods, probabilistic reasoning, or domain-specific constraints to mitigate these challenges, underscoring the complexity of ensuring safe and reliable intelligence in general-purpose systems.

\paragraph{$\bullet$The Halting Problem} It is a decision problem in computability theory \citep{lucas2021origins}, proving that there is no general algorithm capable of deciding whether any arbitrary program halts or runs indefinitely, and it relates closely to other fundamental results (the G\"odel’s incompleteness Theorem \citep{godel1931formal} Theorem, the Chaitin's incompleteness \citep{chaitin1974information} Theorem, and the Rice's Theorem \citep{rice1953classes}). This limitation affects \texttt{BCAI} tasked with verifying program behavior or predicting the termination of dynamic processes. It imposes a theoretical barrier to perfect prediction in self-referential or open-ended systems. The Halting Problem directly impacts the design and limits of \texttt{BCAI}, particularly in areas of reasoning, predictability, and self-analysis.

\paragraph{$\bullet$ Complexity Theory} The unresolved $P$ vs $NP$ problem \citep{cook1971complexity} questions whether every problem whose solution can be verified in polynomial time $NP$ can also be solved in polynomial time $P$. If $P \neq NP$, many critical problems, including optimization and decision-making tasks relevant to \texttt{BCAI}, are computationally intractable. While heuristic and approximate methods can address specific cases, \texttt{BCAI} would face fundamental limits in solving $NP$-complete problems optimally. Complexity theory thus imposes scalability constraints, suggesting that \texttt{BCAI} systems may excel in bounded domains, but struggle with large-scale real-world complexities.

\paragraph{$\bullet$ Shannon's Information Theory} It provides essential tools for optimizing data compression, transmission, and error correction, enabling \texttt{BCAI} to handle vast datasets efficiently \citep{shannon1948mathematical}. However, the Channel Capacity Theorem imposes hard limits on reliable data transmission, particularly in noisy environments. These constraints can hinder \texttt{BCAI} systems reliant on real-time communication or distributed architectures. Additionally, Shannon’s entropy does not distinguish between meaningful signals and noise, leaving \texttt{BCAI} vulnerable to misinformation or adversarial attacks. While supportive in data handling, Information Theory highlights practical barriers to scaling \texttt{BCAI} robustly.

\vspace{\baselineskip}

Let us dwell more on the point that deals with entropy, as it is of special importance in the achievement of \texttt{BCAI}. Shannon's Information Theory relates to cross-entropy \citep{thomas2006elements}, and this is especially relevant because is the basis of many machine learning approaches and variants (e.g. decision trees, neural networks, reinforcement learning, autoencoders, even the famous LLMs). Cross-entropy is used as the loss function during training, which goal is to minimize this loss, which at the same time means reducing the discrepancy between the model predictions and the true distribution. However, there are limits to what cross-entropy minimization can achieve when trying to achieve \texttt{BCAI} behavior. Reducing cross-entropy improves efficiency in information handling, but does not imply that the model has a deep understanding of the concepts it manipulates. A \texttt{BCAI} would require knowledge beyond statistical correlation, implying an ability to model the world and reason about it (e.g. causality \citep{pearl2009causality}), which is not achieved by mere entropy optimization. Moreover, in Information Theory, minimizing entropy is synonymous with finding an efficient encoding of information. For a \texttt{BCAI}, the issue is not only to encode efficiently, but to integrate and use that information to make decisions in varied and unknown environments. This is something that cannot be achieved solely by tuning a model to accurately predict a data set. Finally, entropy and Kolmogorov complexity are related to each other, as the latter measures the minimum length of an object's description. If e.g. an LLM is trained to minimize a cross-entropy loss, it is in a way approximating a more efficient description of its data set. \texttt{BCAI}, however, should be able to understand the underlying ``structure'' of reality, not just find a compact way to describe its observations. 

In summary, Shannon's Information Theory and cross-entropy are crucial for understanding how machine learning works and for evaluating their predictive performance. However, these concepts, while powerful, also highlight the limitations of these models when compared to the requirements for achieving \texttt{BCAI}. Cross-entropy allows measuring how well a model fits the data, but it is not sufficient to capture the depth and breadth of human intelligence, which involves skills such as abstract reasoning, adapting to unprecedented situations, and building a coherent model of the world. Consequently, we may need something beyond this theory to achieve \texttt{BCAI}.

\vspace{\baselineskip}

\begin{figure}[H]
    \centering
    \includegraphics[bb=0 0 600 800,width=0.6\linewidth]{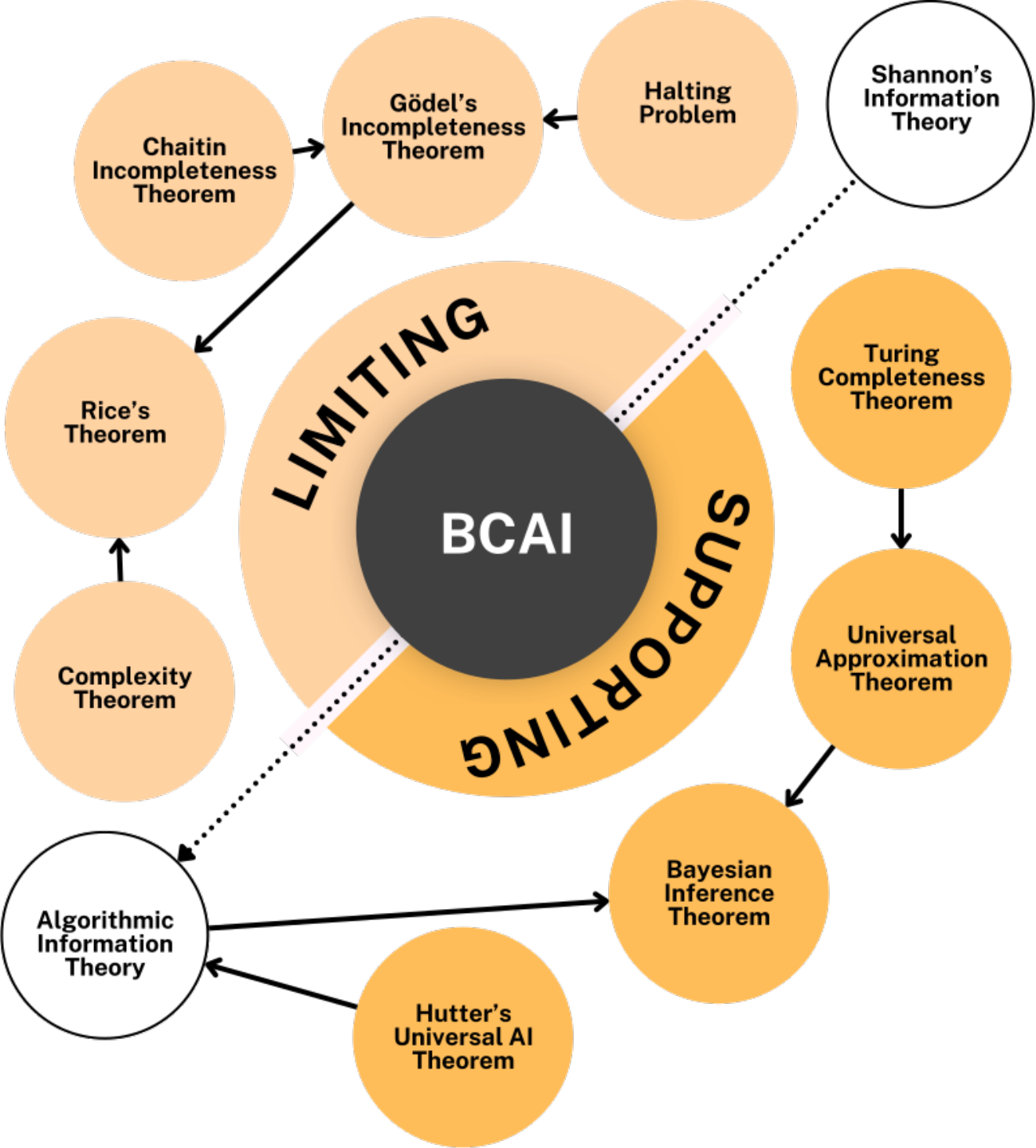}
    \caption[Relationships between theoretical foundations of Computation and AI, and their stance on \texttt{BCAI} feasibility]{Relationships between theoretical foundations of Computation and AI, and their stance on \texttt{BCAI} feasibility. The arrows represent how theorems interact conceptually. Colored nodes highlight enabling principles and emphasize intrinsic computational barriers, and white nodes present ambivalent Computational Intelligence theorems that can either aid or hinder \texttt{BCAI}.}
    \label{fig:BCAI_supp}
\end{figure}

So far our discussions have exposed that the same theorem may support or reject at the same time the possibility to reach \texttt{BCAI}. We have even seen that one theorem can be a limiting factor for another theoretical framework. The chains in Figure \ref{fig:BCAI_supp} represent conceptual or logical relationships between computational theorems, highlighting how their principles interact to either support or limit the feasibility of achieving \texttt{BCAI}. These chains show the flow of ideas and dependencies among the theorems. For example, the main supportive chain (Turing Completeness $\mapsto$ Universal Approximation Theorem $\mapsto$ Bayesian Inference) provides a pathway for \texttt{BCAI} to simulate, learn, and reason across complex tasks. In contrast, the main limiting chain (Halting Problem $\mapsto$ G\"odel's Incompleteness Theorem $\mapsto$ Rice's Theorem) highlights cascading barriers to predictability, reasoning, and self-analysis. Meanwhile, cross-domain insights from Shannon’s Information Theory and Algorithmic Information Theory reveal both the potential for efficient computation and the inherent trade-offs of these approaches. In these ambivalent chains (Shannon’s Information Theory $\mapsto$ Algorithmic Information Theory), we observe that 1) Shannon’s Information Theory enables efficient data compression and error correction, crucial for managing the large datasets needed by \texttt{BCAI}; and 2) that Algorithmic Information Theory extends this by emphasizing minimal data representations and generalization. However, undecidability in finding these optimal representations and limits in data transmission impose constraints on \texttt{BCAI}’s scalability and efficiency. This chain reflects both the enabling and limiting aspects of \texttt{BCAI} development. Finally, we also find mini chains: 

\begin{itemize}
    \item \emph{Limiting:} Complexity Theorem $\mapsto$ Rice's Theorem and Chaitin’s Incompleteness Theorem $\mapsto$ G\"odel’s Incompleteness Theorem. In the first one, the Complexity Theory highlights that many real-world problems are computationally intractable, meaning that \texttt{BCAI} systems may struggle to efficiently solve critical tasks like optimization or decision-making. The Rice’s Theorem adds that even if solutions are found, verifying their correctness or safety is undecidable. Together, these theorems limit \texttt{BCAI}’s scalability and reliability in addressing complex problems and ensuring the quality of its outputs. In the second one, the G\"odel’s Incompleteness Theorem shows that \texttt{BCAI} systems, like any formal system, will encounter true statements they cannot prove, limiting their reasoning capabilities. The Chaitin’s Theorem further emphasizes that some truths require descriptions too complex for the system itself to process. These theorems collectively constrain \texttt{BCAI}’s ability to fully understand, optimize, or explain its own processes.
    \item \emph{Supporting:} Algorithmic Information Theory $\mapsto$ Bayesian Inference and Hutter’s Universal AI $\mapsto$ Algorithmic Information Theory. In the first case, the Algorithmic Information Theory provides a framework for understanding data through its minimal representation, focusing on encoding efficiency and generalization. This aligns directly with Bayesian Inference, where models are updated to fit the data in the most probabilistically efficient way. The Algorithmic Information Theory supports Bayesian reasoning by enabling \texttt{BCAI} systems to prioritize simpler, more generalizable hypotheses, reducing overfitting and improving adaptability. However, the undecidability of finding truly minimal representations limits \texttt{BCAI}’s ability to fully optimize Bayesian models in practice. In the second case, the Hutter’s Universal AI relies on Algorithmic Information Theory to define an ideal decision-making agent. The Algorithmic Information theory underpins this model by focusing on the shortest possible description of environments, allowing the agent to generalize efficiently across tasks. While this highlights the theoretical potential of \texttt{BCAI}, the computational intractability of Solomonoff induction and the undecidability in the Algorithmic Information Theory create practical barriers, making the idealized agent infeasible for real-world implementation.
\end{itemize}

Although some of these approaches may provide a theoretical path that could lead AI to \texttt{BCAI}, there are still issues that can influence its achievement, making the road to \texttt{TAI} a bumpy one e.g., \citep{dobbe2024toward}. As we transition to the next part, we will explore why there are compelling reasons for optimism, as long as \texttt{TAI} is on our side.

\chapter{Opportunities and Perspectives}\label{opps_persp}
\section{Opportunities to Facilitate TAI}\label{tai_yes}

The convergence of technological, financial, and interdisciplinary efforts, combined with the rapid evolution of algorithms and hardware, strongly suggests that the achievement of \texttt{TAI} could not be a question of ``if'', but rather of ``when'' and ``how''. Concerning ``when'', the authors in \citep{gruetzemacher2019forecasting} estimated AI capabilities over the next decade ($2019-2029$), and made questions for forecasting five scenarios of \texttt{TAI}. The same author proposed and outlined in \citep{gruetzemacher2019holistic} a novel framework for forecasting \texttt{TAI}, which integrates a variety of forecasting techniques into a cohesive and holistic approach, aiming to enhance the reliability and comprehensiveness of predictions in this rapidly evolving domain. The famous prediction engine Metaculus\footnote{\url{https://www.metaculus.com/} [Accessed on December 10th, 2024].} also provides forecasts of AI-based technologies, and their impacts on our civilization. 

In this section, we explore how \texttt{TAI} can be realized, offering a brief glimpse into technical and non-technical niches of opportunity -- \emph{green shoots} -- that could pave the way from current AI systems to \texttt{TAI}. We also discuss why certain advances give us reason for optimism about achieving this transformative milestone.

\subsection{Some Technical Green Shoots}

We begin by highlighting several promising technological areas that have the potential to bring us closer to realizing \texttt{TAI}.

\paragraph{$\bullet$ Autonomous Multi-Agent Systems} There is hope for Autonomous Multi-Agent Systems (AMAS), which represent a significant advancement in the field of AI, offering a paradigm where multiple intelligent agents collaborate, compete, or coexist within dynamic environments to achieve individual or collective goals \citep{wang2024survey}. Unlike traditional AI systems that focus on centralized control, AMAS leverage decentralized decision-making, allowing agents to independently perceive their environments, communicate, and adapt their behaviors in real-time. This decentralization enhances scalability, robustness, and flexibility, making AMAS particularly suited for complex, real-world applications such as smart cities, autonomous transportation networks, and distributed robotics. By enabling cooperative problem-solving, AMAS also address challenges that single-agent systems cannot handle efficiently, such as resource allocation, conflict resolution, and emergent behavior. Furthermore, AMAS hold immense potential for developing general-purpose AI systems by enabling coordinated planning and execution of complex tasks across multiple specialized models. Their ability to dynamically collaborate, share knowledge, and adapt to changing environments makes them a powerful framework for tackling challenges that exceed the capabilities of individual systems. As AI research progresses, AMAS are poised to become the next step in AI evolution, bridging the gap between isolated intelligence and truly adaptive, collective intelligence capable of addressing large-scale societal challenges.

\paragraph{$\bullet$ Neural Computation} Our brain emulation is probably the first and the most straightforward strategy that comes to our mind when we think about achieving a superior AI. The earliest work in cybernetics within the emerging field of Neural Computation involved efforts to understand, model, and emulate neurological functions and learning in animal brains. This foundation was established in \citep{mcculloch1943logical}. Neural Computation describes a problem-solving approach that relies on ``learning from experience'' rather than traditional, syntax-driven ``algorithmic'' methods. At its core, it is the study of networks of adaptable nodes (artificial neurons) that, by learning from task examples, store experiential knowledge and make it available for use. Viewed in this way, an artificial neural network is created simply by connecting a set of these adaptable nodes in a structured manner. Lately, the well-known Human Brain Project \footnote{\url{https://www.humanbrainproject.eu/en/} [Accessed on December 10th, 2024].} has come to an end, fostering a profound comprehension of the intricate structure and operations of the human brain, employing a distinctive interdisciplinary approach that bridges the realms of Neuroscience and technology. Many of its advances could certainly lay the groundwork for a future \texttt{BCAI}, relying on new forms of neural computation that transcend the approach that now prevails in modern deep learning models.

Advances in neuromorphic computation and bio-inspired AI architectures are also pushing the boundaries of AI capabilities \citep{pnas2409160121}. These technologies aim to replicate the structure and function of the human brain, making it possible to develop systems that can learn and process information more efficiently, similar to human cognitive processes. For example, companies and research labs are developing neuromorphic chips designed to mimic the way neurons and synapses work, offering a pathway to more flexible and adaptable AI models capable of general intelligence.

\paragraph{$\bullet$ Interactive AI} This is another promising approach \citep{heaven2023generative} which goes one step further than conversational AI; AI systems that can autonomously execute tasks, collaborate with other technologies, and handle multi-step workflows, some of them derived from AMAS. Interactive AI would enhance user interaction by integrating decision-making, planning, and task management, moving beyond content generation to practical applications that actively solve problems across diverse fields. This transition marks a more complex and action-oriented phase of AI development.

\paragraph{$\bullet$ Computational Power and Specialized Hardware} The growth of computational power continues to follow an exponential trend, supported by the development of specialized AI hardware such as Graphics Processing Units (GPUs), Tensor Processing Units (TPUs), and emerging quantum computing technologies. These advancements enable the training of larger, more sophisticated models that can handle tasks requiring massive parallel processing and optimization \citep{garisto2024cutting}. As computational capacity increases, it is likely to bridge the gap necessary for achieving \texttt{BCAI}.

\paragraph{$\bullet$ Self-Improving and Self-Learning Algorithms} AI systems that leverage reinforcement learning and unsupervised learning have shown the ability to self-improve without constant human input. Techniques like meta-learning, where AI systems learn how to learn, indicate that algorithms are moving toward autonomous development. For instance, Generative Pre-trained Transformers (GPT) \citep{yenduri2024gpt} have progressively become more efficient in learning from vast datasets, suggesting that future models could independently refine their learning processes, mimicking aspects of human cognition, or autonomy \citep{firat2023if,wu2023autogen}.

Building on this trajectory, \citep{clune2019ai} proposed a paradigm shift, suggesting that rather than directly programming AI systems, researchers could focus on developing AI-generating algorithms (AI-GAs). They would harness principles of artificial evolution, self-improvement, and open-ended learning to autonomously evolve increasingly sophisticated AI systems. The work argues that this approach may offer a more scalable and adaptable route to \texttt{BCAI} than some methods. By leveraging evolutionary principles, AI-GAs could progressively develop AI systems with general intelligence capabilities. Approaches like \citep{real2020automl} are focused on the same idea: machine learning algorithms are evolved from scratch using evolutionary computation, i.e. a system that starts with basic mathematical operations and evolves increasingly complex algorithms through a process of mutation, crossover, and selection. These evolutionary paradigms, which have been present in recent achievements in general-purpose models \citep{poyatos2024evolutionary}, open up new possibilities for automatically generating effective ML algorithms in an unsupervised manner, bypassing the need for manual design.

\paragraph{$\bullet$ Generalization Capabilities} AI systems are becoming increasingly capable of generalizing across domains, even under data frugality (the so-called \emph{few-shot} and \emph{zero-shot} learning regimes) \citep{song2023comprehensive,wang2019survey}. Models like AlphaZero \citep{zhang2020alphazero}, which mastered chess, Go, and Shogi using the same framework without domain-specific adjustments, highlight AI’s ability to adapt knowledge and strategies across different fields. This kind of cross-domain adaptability is a core requirement for \texttt{TAI} and indicates that current models are laying the groundwork for systems with broader, general capabilities.

\paragraph{$\bullet$ Open-World Learning} It is a critical paradigm in advancing AI systems toward adaptability and robustness in dynamic, real-world environments. Unlike traditional machine learning models, which operate under closed-world assumptions with predefined classes, these approaches acknowledge the existence of unknown or novel categories during deployment. \textit{Open-Set Recognition} focuses on detecting and handling inputs that do not belong to the training set, ensuring the model can differentiate between known and unknown classes. \textit{Open-World Learning} extends this capability by not only recognizing unknowns, but also dynamically incorporating them into the learning process, enabling continuous adaptation (\textit{Continual Learning} \citep{wang2024comprehensive}). These paradigms are pivotal for applications like autonomous vehicles, security systems, and robotics, where encountering unfamiliar scenarios is inevitable. Recent advances in uncertainty quantification, novelty detection, and few-shot learning have significantly enhanced these approaches, yet challenges remain in achieving scalability, minimizing false positives, and ensuring reliable consolidation and integration of new knowledge into the model. Together, \textit{Open-Set Recognition} and \textit{Open-World Learning} push AI systems closer to human-like adaptability and lifelong learning capabilities, being capable of \emph{managing the unknown} \citep{barcina2024managing}. 

\paragraph{$\bullet$ Sophisticated Simulations and Virtual Environments} AI models are increasingly trained in highly sophisticated virtual environments that simulate complex, real-world scenarios. These simulations allow AI systems to practice and develop a wide range of skills and strategies autonomously, without human intervention, to the extent of learning internal representations of the simulation itself (the aforementioned \emph{world models}). Some AI models have been trained in virtual environments that tested strategic thinking and adaptability, showing that AI systems can acquire complex skills through simulation. This technology is crucial for developing a \texttt{BCAI} that can operate and learn in diverse, unpredictable environments.

\paragraph{$\bullet$ Causality and Reasoning} Unlike traditional machine learning models, which primarily rely on statistical correlations, causality \citep{pearl2009causality} aims to uncover the underlying mechanisms that govern real-world phenomena. By integrating causal reasoning into AI systems, researchers hope to address key limitations of current approaches and pave the way for achieving the level of understanding and adaptability required for \texttt{BCAI}.

Causal reasoning and counterfactual thinking \citep{gerstenberg2024counterfactual} are deeply interconnected concepts that play a fundamental role in advancing AI toward \texttt{BCAI}. Causal reasoning focuses on identifying and understanding cause-and-effect relationships, enabling AI systems to move beyond statistical correlations to uncover the mechanisms driving observed phenomena. Counterfactual thinking builds on this foundation by allowing AI to evaluate ``what-if'' scenarios, imagining alternative outcomes based on hypothetical changes to causal factors. Together, these capabilities empower AI to not only predict future events with greater accuracy, but also to generalize knowledge across diverse domains and contexts. By successfully integrating these two dimensions, researchers aim to develop AI systems that can reason abstractly, adapt to new and unseen situations, and make decisions grounded in a deep understanding of the world’s causal structure, all of which are essential for achieving \texttt{BCAI}.

Despite its potential, the integration of causality into AI systems faces significant challenges \citep{bishop2021artificial}. These include the computational complexity of causal discovery, the difficulty of acquiring sufficient data for causal inference, and the need for interdisciplinary collaboration to develop robust causal frameworks. Nevertheless, ongoing advancements in causal machine learning offer promising directions \citep{liu2024large,weinberg2024causality} for overcoming these hurdles.

\paragraph{$\bullet$ Quantum Computing} The integration of quantum computing into AI research holds the potential to significantly advance the development of \texttt{TAI}. In some cases quantum computing can offer computational capabilities that surpass classical systems, enabling more efficient processing of complex algorithms and large datasets, which are essential for sophisticated AI applications. 

One of the primary advantages of quantum computing in AI is its ability to handle high-dimensional data spaces and perform complex probability distributions more efficiently than classical computers. This capability is particularly beneficial in machine learning tasks, where quantum algorithms can provide exponential speedups in training models and solving optimization problems that lie at the heart of their learning algorithms \citep{nguyen2024machine}. Moreover, quantum computing facilitates the development of new AI algorithms that leverage quantum principles, leading to more powerful and efficient AI systems. For instance, quantum-enhanced machine learning algorithms can outperform their classical counterparts in specific tasks, thereby accelerating the progress toward more advanced AI capabilities \citep{Klusch2024}.

In the specific case of causal reasoning, quantum computing holds significant potential to revolutionize this field by addressing its computational challenges and enabling more sophisticated analyses \citep{barrett2019quantum,hutter2023quantifying}. The exploration of causal structures often involves traversing an exponentially large space of possible models, a task that is computationally prohibitive for classical systems. Quantum algorithms, such as the Quantum Approximate Optimization Algorithm (QAOA) and Grover's search, can accelerate this process by exploring multiple causal hypotheses in parallel, thereby reducing the time required for causal discovery. Additionally, quantum-enhanced machine learning can facilitate more efficient estimation of causal effects in high-dimensional datasets by leveraging quantum speedups in optimization and linear algebra. Moreover, the ability of quantum systems to simulate complex probabilistic relationships allows for the modeling of counterfactual scenarios with greater accuracy, a cornerstone of causal inference. By combining these capabilities, quantum computing could significantly enhance our understanding of causal relationships, particularly in dynamic and high-dimensional systems, paving the way for breakthroughs in fields ranging from medicine to social sciences.

However, the integration of quantum computing into AI also presents difficulties in its realization, such as the need for specialized hardware and the development of new programming paradigms. Addressing these challenges requires interdisciplinary collaboration and continued research to fully realize the potential of quantum computing in advancing transformative AI.

\subsection{Some Non-Technical Green Shoots}

As we have observed, the path to achieving \texttt{TAI} is not limited to technical advancements, but also includes other non-technical opportunities that may be equally significant.

\paragraph{$\bullet$ Integration of Interdisciplinary Approaches and Global Collaboration} The pursuit of \texttt{TAI} is not limited to one field; it integrates Neuroscience, Cognitive Science, Computer Science, Mathemathics, and even Sociology and Philosophy, among others. This interdisciplinary approach, coupled with global collaboration across academia, industry, and governments, creates a fertile environment for rapid advancements and synergistic public-private partnerships that can help align research objectives with societal values and industrial goals.

\paragraph{$\bullet$ Significant Investment and Economic Incentives} There is unprecedented financial investment in AI development from technological giants, as well as from governments recognizing AI as a strategic technology for their nations. The economic and strategic incentives behind achieving \texttt{TAI} are enormous, driving the allocation of vast resources and talent into AI research. These investments ensure continuous progress and innovation, increasing the likelihood of \texttt{TAI}’s development.

\paragraph{$\bullet$ Massive Data Availability and Enhanced Data Processing Techniques} The availability of big data continues to grow, providing a vast amount of information for AI systems to learn from (we hereby renew the call for public data to truly become an opportunity). The combination of these extensive datasets with enhanced data processing techniques (e.g., federated learning, transfer learning) allows AI to train on diverse and comprehensive sources of information. This availability of high-quality data are crucial for building systems that are capable of understanding and interacting with the world in a general and adaptable manner.

\paragraph{$\bullet$ Growing Focus on Ethical AI and Governance Frameworks} The increasing focus on developing ethical and regulatory frameworks around AI indicates that societies and governments are preparing for the arrival of \texttt{TAI}. Efforts such as the \textit{EU’s AI Act}, the mentioned \textit{AB 3030} in California, or the \textit{OECD’s AI Principles}, or the NIST framework\footnote{\url{https://www.nist.gov/itl/ai-risk-management-framework}}, demonstrate that there is a proactive movement to manage and integrate advanced AI technologies responsibly. The establishment of these governance models not only ensures a controlled and ethical development path but also attracts more researchers and investors to work on achieving \texttt{TAI} in a way that aligns with societal values.

\vspace{\baselineskip}

We will now highlight, in the opinion of the authors and other researchers, what could be the most promising \textit{green shoot} among those currently available.
\section{A Great Hope for TAI: the ``Science Explosion''}\label{the_hope}

We can find some literature with experts' predictions on the pace of AI progress, the nature and impacts of advanced AI systems, and their own views on the \emph{green shoots} for the future development of this technology. In \citep{grace2024thousands}, the authors examine a particularly intriguing aspect of the survey, where AI researchers are asked to provide their predictions. A notable point of divergence arises from respondents who anticipate significant disruption to scientific progress due to AI advancements. At this juncture, we pause to explore a \emph{green shoot} that, in our belief, could catalyze the emergence of \texttt{TAI}: AI serving as the impetus for a \emph{scientific revolution}. However, it appears that we are not currently on the right track.

In the opinion of some experts, scientific progress is slowing down \citep{cowen2019rate,wu2019large,macher2024there}. The number of research papers in science and technology has surged dramatically in recent decades \citep{park2023papers}; however, an analysis of their content reveals a decline in their ``disruptiveness'', a measure of how significantly these papers diverge from and challenge existing literature. This suggests that while the volume of research is increasing, the extent to which new studies introduce groundbreaking ideas or shift established paradigms has diminished. Another factor contributing to this decline is the emphasis on quantitative metrics, such as citation counts, which can encourage researchers to prioritize publication volume over innovative research. In this direction, we highlight the Declaration on Research Assessment (DORA)\footnote{\url{https://sfdora.org/} [Accessed on December 10th, 2024].} as a pivotal initiative in modern academia, advocating for more equitable, diverse, and meaningful evaluation of research outputs. It seeks to align assessment practices with the principles of open science, transparency, and societal relevance, ensuring that researchers are evaluated on the true impact of their work. Despite the clear motivation and solid rationale for the initiative, DORA faces two main challenges for its widespread adoption: 1) the resistance due to entrenched reliance on journal metrics, and 2) the difficulty in implementing new evaluation frameworks across disciplines. This environment may discourage risk-taking, as academics and institutions often favor predictable results that align with existing theories.

However, technology differs fundamentally from science: it represents the application of scientific discoveries rather than the discovery process itself. Science is the systematic exploration of the physical and natural world, involving observation, experimentation, and the validation of theories against gathered evidence. When we perceive e.g., our latest smartphone as a symbol of ``unstoppable progress'', what we are really highlighting is not a new scientific breakthrough, but rather the industrial-scale refinement of technologies built on well-established scientific principles.

Therefore, in order to stop this slowdown in scientific development, AI could play an expanding role in science, encompassing a wide range of fields, serving both as a catalyst for scientific breakthroughs and as a crucial tool in the research process \citep{hou2023using,wang2023scientific,agrawal2024artificial,lawrence2024accelerating}. This would mark the beginning of a new era characterized by accelerated discoveries, enabling progress at the frontiers of science and achieving results beyond the capabilities of existing methodologies. Such acceleration would have the potential to address critical societal challenges, including climate change, public health, and the green and digital transitions, among many others. One of the best examples of this opportunity is the 2024 chemistry Nobel-prize-winning \textit{AlphaFold} \citep{jumper2021highly}, with the capacity of combating diseases or addressing plastic pollution, and with a remarkable impact on some of our most pressing global challenges. Its database will support the scientific community in their research, and unlock entirely new pathways for scientific discovery. As another notable recent example, it is worth highlighting the study of \cite{hou2024using}, where the \textit{LucaProt} AI identified $161,979$ putative RNA virus species and $180$ RNA virus supergroups. This study sets a new benchmark for computational tools in virus discovery, highlighting the vast untapped diversity of the global RNA virome. Currently, many expectations \citep{ruoss2024grandmaster} are being devoted to LLMs and their architecture based on Transformers, i.e. a type of neural network architecture introduced in \citep{vaswani2017attention}. Finally, we would like to bring the focus on \cite{kudiabor2024virtual}, where in a virtual laboratory coexist several ``AI scientists'' (i.e. LLMs with defined scientific roles) that collaborate to achieve objectives planned by human researchers. However, it remains unclear whether in all these cases their mathematical reasoning capabilities have genuinely advanced, so perhaps we should resolve it before assigning AI an expanding role in science \citep{mirzadeh2024gsm}. 

In developing an efficient and impactful policy on AI for science, we could focus on two key directions:

\paragraph{$\bullet$ Accelerating the adoption of AI by scientists} This involves creating essential enablers to facilitate the integration of AI into scientific research. These enablers include improving access to high-quality data, ensuring sufficient computational power, and nurturing talent within the AI field. By providing these resources, we aim to empower scientists to fully leverage AI tools, driving innovation and progress across diverse research areas.
    
\paragraph{$\bullet$ Monitoring and steering the impact of AI on the scientific process} This direction focuses on addressing the specific challenges that arise from the integration of AI in science, such as maintaining scientific integrity and upholding methodological rigor. It involves closely overseeing how AI influences the research process, ensuring that AI applications in science are transparent, reproducible, and adhere to the highest standards of scientific practice. This approach aims to safeguard the reliability of AI-driven research and to guide its responsible use in advancing scientific knowledge.

\vspace{\baselineskip}

Even if we were to see \texttt{TAI} become a reality, we would need a new way of approaching and understanding our new world. Maybe \texttt{TAI} would emerge but we would not be quite ready to have it in our midst?.
\section[Would We Need a New Ethical and Philosophical Perspective for TAI?]{Would We Need a New Ethical and Philosophical Perspective for TAI?}\label{new_rel}

So far, we have explored how \texttt{TAI} could be achieved in the future. But once we reach that point, we must ask ourselves: \emph{what comes next?} A world with \texttt{TAI} could bring profound changes, particularly to the ethical and philosophical frameworks guiding our experiences and interaction with this disruptive technology. As we have discussed throughout this article, we must also consider how an AI-driven transformation of our civilization could reshape our perspectives on ethics and philosophy.

Indeed, the advent of \texttt{TAI} could compel society to confront fundamental ontological and ethical challenges, potentially necessitating the emergence of a new philosophical framework. From an ontological perspective, \texttt{TAI} could radically challenge existing conceptions of consciousness and identity. Functionalist theorists \citep{putnam1967psychological,schneider2017daniel} argue that consciousness is a function of processes rather than a substrate. If AI systems achieve a form of self-reflective understanding, the traditional Cartesian dualism that separates mind from body may require reinterpretation to account for non-biological entities. The definition of ``personhood'' could similarly expand, following ethical frameworks posited in \citep{singer2016practical}, where it is suggested that ``personhood'' is defined by a level of rationality and self-awareness, rather than by biological origin.

Ethically, the question arises as to whether AI could become a moral agent. The philosopher of the Theory of Responsibility \citep{jonas1979imperative} posits that those with power must bear moral obligations toward all forms of life impacted by their actions. An AI with transformative power might bear a new kind of responsibility, not only toward humans but also toward its own existential condition. Additionally, utilitarianism \citep{mill1863utilitarianism}, which evaluates morality based on the consequences of actions, could be expanded to include non-human entities, requiring a recalibration of our ethical calculations.

The ``transhumanist'' movement \citep{bostrom2014superintelligence} considers AI as a logical step in human evolution and warns of existential risks if alignment with human values fails. The author of \citep{harari2017homo} has posited that a new ``religion of artificial intelligence'' could emerge, one that attributes divinity or quasi-divine status to these entities, potentially merging or contending with existing religious systems. These currents suggest a potential need for a ``post-human” philosophy, addressing humanity’s place alongside entities of superior intelligence. Thus, in the context of these transformative changes, a ``meta-religion'' or ``meta-philosophy'' could emerge, oriented toward reconciling humanity with the presence of synthetic intelligences that redefine traditional questions of purpose and morality.

In conclusion, \texttt{TAI} challenges the very foundation of human-centric ontology, ethics, and spirituality. Addressing these profound questions may necessitate a philosophical -- and perhaps, even a spiritual framework -- that embraces coexistence with intelligent non-biological entities, while safeguarding human dignity and well-being in this new reality.
\chapter{Reflections and Conclusions}\label{ref_conc}
\section{Discussion and Open Thoughts from the Community}\label{disc}

What better way to address \texttt{TAI} conclusively than by drawing on the insights of some of the greatest visionaries and scientists in AI? These famous quotes are deeply intertwined with the concept of \texttt{TAI}, though not all of them align perfectly. \texttt{TAI} is a highly subjective goal, and by examining these quotes, we aim to offer a profound and comprehensive discussion that enriches the conclusions of this article.

\begin{quote}
    ``\textit{The saddest aspect of life right now is that science gathers knowledge faster than society gathers wisdom.}''
    \\
    --- \textit{Isaac Asimov}
\end{quote}

The misalignment between technological progress and societal readiness can lead to unintended consequences, as AI's capabilities evolve faster than our collective wisdom to manage its risks and ensure that its deployment aligns with human values and long-term well-being. Thus, while \texttt{TAI} holds tremendous promise for progress, it also underscores the urgent need for deeper societal reflection and a more deliberate pace in crafting ethical standards and policies that can guide these advancements responsibly.

\begin{quote}
    ``\textit{I do not fear computers. I fear the lack of them.}''
    \\
    --- \textit{Isaac Asimov}
\end{quote}

The fear of a ``lack'' of such technologies underscores the concern that, without AI, humanity might miss crucial opportunities to enhance global well-being, optimize resources, and make faster scientific discoveries. Yet, this enthusiasm must be balanced with a thoughtful approach to implementation. As society becomes increasingly dependent on AI for progress, it is crucial to develop the necessary ethical and regulatory frameworks to ensure these systems are used responsibly. Thus, while the absence of AI could limit humanity’s potential to tackle pressing global issues, the challenge remains in ensuring that the presence of AI aligns with societal values and long-term goals.

\begin{quote}
    ``\textit{It is change, continuing change, inevitable change, that is the dominant factor in society today. No sensible decision can be made any longer without taking into account not only the world as it is, but the world as it will be.}''
    \\
    --- \textit{Isaac Asimov}
\end{quote}

\texttt{TAI} represents a profound shift in the technological landscape, one that continually reshapes our understanding of what is possible. As AI systems advance, they bring rapid changes that influence various sectors, from industry and education to healthcare and governance. These developments challenge policymakers, researchers, and society to make decisions that are forward-looking, recognizing the dynamic and evolving nature of AI. Ignoring the future potential of AI or failing to anticipate its long-term effects could lead to missed opportunities or unintended consequences. Thus, embracing AI as a catalyst for change requires a vision that understands its current capabilities and envisions how these technologies will shape the world to come. This perspective underscores the need for proactive and adaptive approaches, ensuring that the societal integration of AI remains aligned with our evolving aspirations and challenges.

\begin{quote}
    ``\textit{The advance of civilization is nothing but an exercise in the limiting of privacy.}''
    \\
    --- \textit{Isaac Asimov}
\end{quote}

As AI continues to develop and integrate into various aspects of society, its transformative potential often comes at the cost of personal privacy. Advanced AI systems, particularly those involved in data analysis, surveillance, and predictive modeling, rely heavily on vast amounts of personal and behavioral data to function effectively. While these systems can drive innovations in healthcare, public safety, and personalized services, they also raise significant concerns about how much privacy individuals must sacrifice for these benefits. The collection and processing of data by AI challenge traditional notions of privacy, as increasingly detailed information about individuals is required to fuel the progress of these technologies. This tension highlights a central dilemma in the era of a possible \texttt{TAI}: the balance between the societal benefits of AI-driven advancements and the need to protect personal freedoms. As AI reshapes the world, it is essential to question how much of our privacy we are willing to concede in the name of civilizational progress, and to ensure that appropriate safeguards and ethical guidelines are in place to protect individual rights.

\begin{quote}
    ``\textit{I think that the progress in AI is going to continue, and the only way to get the benefits of AI is to make it safe.}''
    \\
    --- \textit{Geoffrey Hinton}
\end{quote}

AI evolves rapidly, and its potential to reshape industries, advance scientific research, and address global challenges is undeniable. However, this progress must be accompanied by a strong emphasis on safety, ensuring that the deployment of AI systems does not introduce new risks or exacerbate existing ones. Ensuring a safe AI involves establishing robust ethical guidelines, implementing transparency in AI processes, and ensuring that these technologies are aligned with human values. Without these safeguards, the transformative power of AI could lead to unintended consequences, such as biased decision-making, privacy violations, or even threats to social stability. Therefore, to fully realize the promise of AI, it is crucial to integrate safety considerations into its development, allowing society to harness its benefits while minimizing risks. This balance between innovation and responsibility is key to ensuring that AI serves as a positive force for progress.

\begin{quote}
    ``\textit{If we allow ourselves to be enchanted by the possibilities of machines, we risk diminishing the importance of human creativity and agency.}''
    \\
    --- \textit{Jaron Lanier}
\end{quote}

There is a growing fascination with the AI's ability to automate tasks, generate insights, and even engage in creative processes. While these capabilities can greatly enhance productivity and open new avenues for innovation, they also pose a risk of overshadowing the unique contributions that human creativity and decision-making bring to various fields. The allure of AI's efficiency might lead us to rely too heavily on its outputs, potentially suppressing human imagination and reducing our role to mere operators of complex systems (the \emph{irrelevance of humans} as a big danger \citep{harari201821}). In the pursuit of leveraging AI's full potential, it is essential to maintain a balance that ensures AI complements rather than replaces human agency. True progress in AI should involve using these technologies to augment human capabilities, allowing individuals to focus on creativity, ethical considerations, and complex problem-solving that cannot be entirely captured by algorithms. This perspective emphasizes the need for thoughtful integration of AI, where human values and ingenuity remain central to its application, ensuring that we do not lose sight of the irreplaceable aspects of human intellect in our fascination with technological progress.

\begin{quote}
    ``\textit{The world of the future will be an ever more demanding struggle against the limitations of our intelligence, not a comfortable hammock in which we can lie down to be waited upon by our robot slaves.}''
    \\
    --- \textit{Norbert Wiener}
\end{quote}

This perspective emphasizes that, despite the advancements in AI and automation, the future will require continuous effort and intellectual engagement from humanity. While AI has the potential to automate many tasks and alleviate certain burdens, it does not imply that humans can become passive observers. Instead, the presence of advanced AI systems demands that we constantly push the boundaries of our understanding, adapting to new challenges and ensuring that AI serves human interests responsibly. The real struggle lies in managing AI's complexities, aligning it with societal goals, and ensuring that its deployment enhances rather than diminishes human agency. This ongoing effort to guide and integrate AI into the fabric of society underscores the importance of wisdom, foresight, and active stewardship as we navigate the technological landscape of the future. Remarkably, this view does not relegate the human to irrelevance, contrarily to Yuval Noah Harari in \citep{harari201821}.

\begin{quote}
    ``\textit{AI is not, by itself, a threat. It is our use of AI that presents potential risks.}''
    \\
    --- \textit{John McCarthy}
\end{quote}

McCarthy's perspective highlights that the inherent capabilities of AI, no matter how advanced, do not pose a danger in isolation. Rather, the true challenge lies in how humans choose to develop, deploy, and govern these technologies. \texttt{TAI} holds immense potential to address global challenges, enhance productivity, and drive scientific innovation. However, without careful oversight and ethical guidelines, it can also lead to unintended consequences, such as biases in decision-making, erosion of privacy, or misuse in harmful applications. Therefore, the key to realizing the positive impact of AI while mitigating its risks is to ensure that its use is guided by well-thought-out principles, transparency, and a commitment to societal well-being. This approach places the responsibility squarely on human actors to shape AI in ways that align with our values and long-term interests, ensuring that this powerful tool contributes positively to the future of humanity.

\begin{quote}
    ``\textit{If a human can't understand how an AI system works, then the AI system should not be deployed.}''
    \\
    --- \textit{Yoshua Bengio}
\end{quote}

Bengio's view emphasizes the importance of transparency and interpretability in the development and deployment of AI systems. While \texttt{TAI} has the potential to revolutionize all fields, its impact can only be truly positive if humans can understand and trust the decisions made by these systems. When AI operates as a ``black box'', with complex algorithms that are opaque even to their creators, it becomes difficult to assess their fairness, reliability, and potential biases. This lack of understanding can undermine trust and lead to unintended consequences, especially in critical applications that directly affect people's lives. Therefore, achieving the promise of \texttt{TAI} requires a commitment to creating systems that are not only powerful, but also transparent and explainable. By ensuring that AI systems are understandable to those who use them, we can maintain human oversight, build public trust, and ensure that AI is deployed in a way that aligns with ethical principles and societal values.

\begin{quote}
    ``\textit{We need to think about the future of AI not as something that just happens to us but as something that we actively shape.}''
    \\
    --- \textit{Yoshua Bengio}
\end{quote}

This second perspective from Y. Bengio underscores the idea that AI should not be viewed as an inevitable force of change that passively transforms society, but rather as a powerful tool whose direction and impact are determined by human choices. As AI technologies continue to advance, their effects on society, economy, and culture will depend largely on the values and priorities that guide their development. By actively shaping the future of AI, we can ensure that it aligns with ethical principles, promotes social equity, and addresses global challenges. This requires thoughtful governance, public engagement, and a commitment to using AI in ways that enhance human well-being. Rather than allowing AI to dictate the terms of change, we must take a proactive role in steering its course, ensuring that it serves as a positive force for innovation and progress. In doing so, we can create a future where AI supports, rather than disrupts, the values and aspirations of our society.

\begin{quote}
    ``\textit{AI is about extending our minds, rather than replacing them.}''
    \\
    --- \textit{Demis Hassabis}
\end{quote}

Hassabis' opinion emphasizes that the true potential of AI lies not in substituting human intelligence, but in augmenting and expanding our cognitive capabilities. \texttt{TAI} offers opportunities to enhance problem-solving, creativity, and decision-making by providing new insights and handling complex data analysis at a scale beyond human capacity. Rather than viewing AI as a competitor to human intelligence, this approach envisions a collaborative relationship where AI acts as a tool that empowers individuals to achieve more, unlocking new possibilities in fields like science, medicine, and arts. By leveraging AI as an extension of human thought, we can push the boundaries of what we can understand and create, allowing AI to complement our strengths while preserving the uniquely human elements of empathy, judgment, and creativity. In this way, the development of AI can be guided toward a future where it enhances human potential and supports our aspirations, rather than diminishing our role in shaping the world. Notably, this view is also shared by Asimov's belief.

\begin{quote}
    ``\textit{The development of AI is not a race, it’s a journey, and the most important thing is to ensure that everyone benefits.}''
    \\
    --- \textit{Fei-Fei Li}
\end{quote}

She highlights that the advancement of AI should be approached with patience and careful consideration, rather than with a competitive mindset focused solely on speed. The true promise of AI lies in its potential to improve lives and address global challenges, but achieving these goals requires a thoughtful, inclusive approach. By treating AI development as a journey, we can focus on building systems that are safe, ethical, and aligned with societal values, ensuring that their benefits reach all segments of society rather than just a few. This means prioritizing fairness, transparency, and access, and working to bridge the gaps between technological innovation and social equity. In doing so, the development of AI becomes a collective effort aimed at fostering positive change, where the ultimate goal is not merely to be the first, but to ensure that the transformative power of AI serves the greater good and contributes to a more just and equitable world.

\begin{quote}
    ``\textit{No matter how sophisticated the computation is, how fast the CPU is, or how great the storage of the computing machine is, there remains an unbridgeable gap (a ``humanity gap'') between the engineered problem solving ability of machine and the general problem solving ability of man.}''
    \\
    --- \textit{J. Mark Bishop}
\end{quote}

The phrase underscores a fundamental philosophical and practical limitation in AI, a viewpoint shared by numerous researchers and thinkers in the field. This gap signifies more than just a difference in processing speed or efficiency; it reflects an essential distinction in how problems are approached and understood by humans versus machines. Humans possess a deeply contextual, adaptive, and flexible cognitive framework that allows for nuanced understanding, abstract reasoning, and generalization across vastly different types of problems. This ability to approach problems with insight, intuition, and emotional or ethical judgment, is something that engineered systems, no matter how advanced, inherently cannot provide. AI systems, by design, operate within the constraints of their programming, relying on structured data, patterns, and predefined algorithms. They lack the subjective, experiential knowledge and adaptability that characterize human intelligence.

\vspace{\baselineskip}

Despite their disparate perspectives, the majority of these thinkers and scientists concur that AI has the potential to be a transformative force for humanity. However, there is a divergence of opinion as to the extent of the caution that should be exercised, and the manner in which the integration of AI into society should be managed. While some, such as McCarthy and Asimov, evince a more pronounced technological optimism, others, including Hinton, Lanier, and Wiener, emphasize the necessity of meticulously addressing the risks and ethical implications, particularly regarding the control and transparency of the technology. Regardless of divergent views, the shared perspective is that the responsibility for guiding the development of AI rests with humanity, and it is up to us to determine how we direct its impact in the future.
\section{Conclusion}\label{conc}

Regardless of the role that AI ultimately assumes within our civilization, we still hold the reins, with the opportunity to harness its full potential and ensure that it becomes a positive turning point in nearly every aspect of our society. By actively shaping the direction of AI development, prioritizing ethical considerations, and fostering inclusive access to its benefits, we can guide this technology to address complex challenges and enhance human well-being. The future of AI remains in our hands, allowing us to channel its transformative capabilities toward progress and improvement across diverse fields. We have to do everything we can to take advantage of AI development in the right direction, or else we will be confirming some assumptions that we would not like to \citep{garrett2024artificial}. By unveiling and dissecting transformative AI, this work aims to contribute to the broader discourse on AI's long-term implications and its pivotal role in potentially reshaping the future of our civilization.

We conclude by offering our own perspective on the transformative role of Artificial Intelligence, which aligns closely with the implicit consensus found in the perspectives reviewed earlier:

\begin{quote}
    ``\textit{Let us start taking real control on Artificial Intelligence, as it seems we still have time.}''
    \\
    --- \textit{The authors}
\end{quote}

\chapter*{Acknowledgements}

We would like to thank our colleague Dr. Sergio Gil L\'opez for the interesting discussions held on some of the topics covered in this work.

\nextpage

% After the \backmatter command, sections will not be numbered.
\backmatter

\chapter{About the Authors}
\begin{spacing}{1.0}
\begin{minipage}[t]{0.3\textwidth}
  \centering\raisebox{\dimexpr \topskip-\height}{%
  \includegraphics[width=\textwidth]{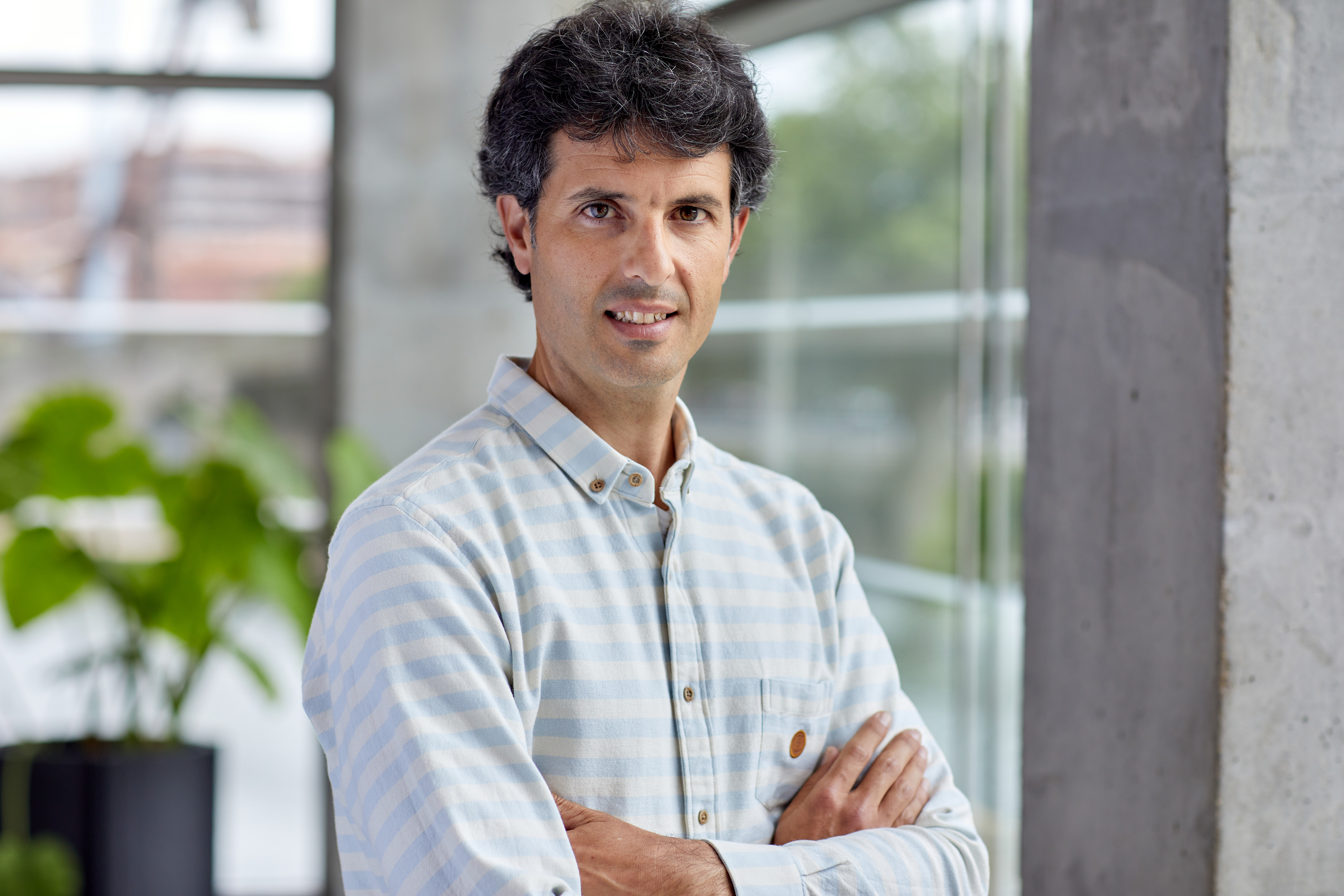}} % Foto del autor
\end{minipage}
\hfill
\begin{minipage}[t]{0.65\textwidth} % Ancho para el texto del autor
    \textbf{Jesus L. Lobo, PhD} \\
    He is a passionate of AI with extensive experience as a Research Scientist at \emph{Tecnalia}\footnote{\label{tecn}\url{https://www.tecnalia.com/en/}}, the largest applied research center in Spain and one of the most relevant in Europe. His mission is to explore, develop, and transfer scientific-technological solutions in AI that generate value for society and organizations. His field of expertise gravitates on \textit{Adaptive AI}, which addresses the challenges posed by dynamic and changing environments for machine learning systems. He is also interested in \textit{AI Ethics}, \textit{AI Governance}, and \textit{AI Alignment} with human values, among other topics. Finally, he has participated in several research and innovation projects, published several scientific papers in high impact journals and conferences, and contributed to the dissemination of AI nationally and internationally.
\end{minipage}

\vspace{0.7cm} % Espacio entre autores

\noindent\begin{minipage}[t]{0.3\textwidth}
    \centering\raisebox{\dimexpr \topskip-\height}{%
    \includegraphics[width=\textwidth]{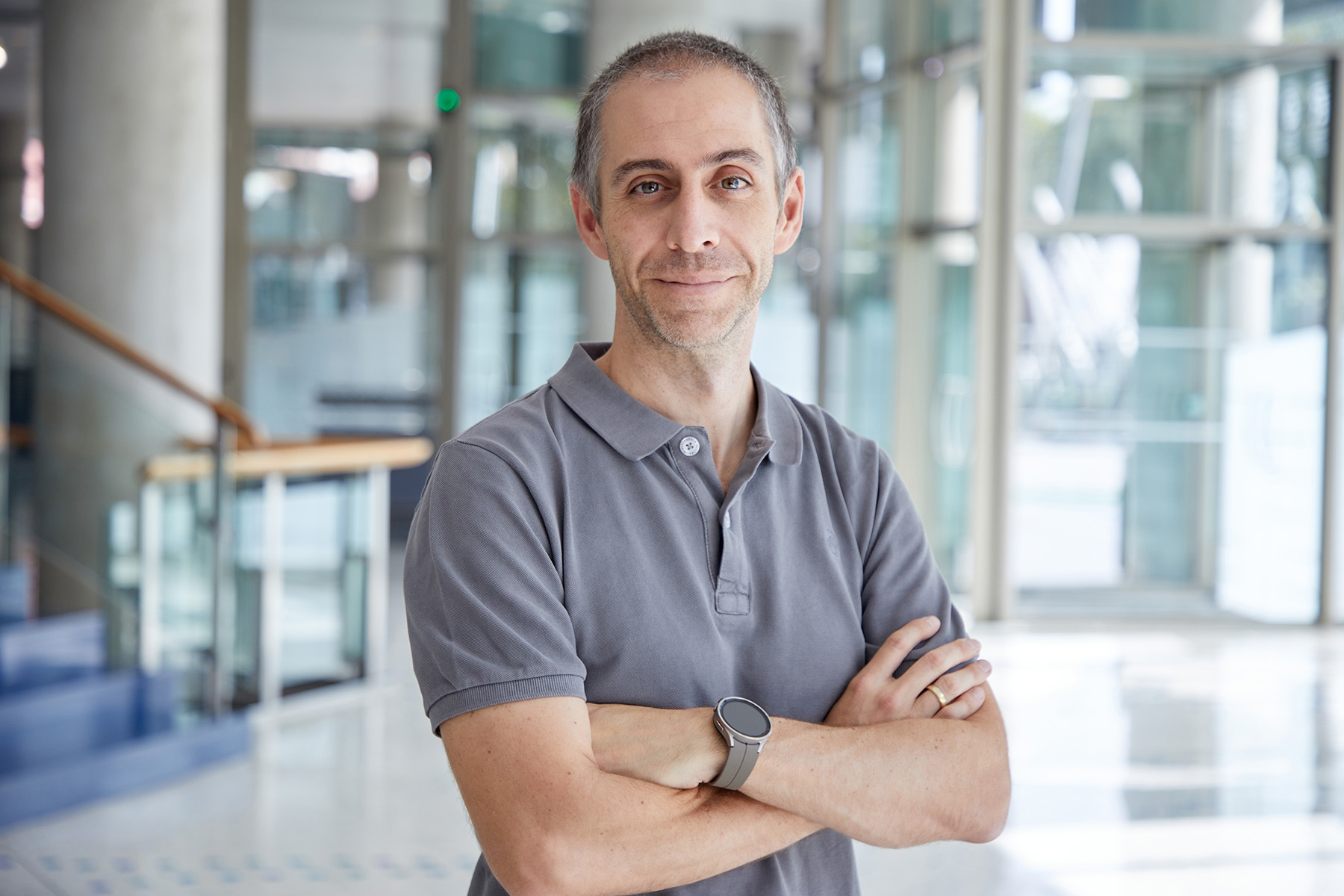}}
\end{minipage}
\hfill
\begin{minipage}[t]{0.65\textwidth}
    \textbf{Javier Del Ser, PhD} \\
    He is the Chief Scientific \& Technology Head of AI at \emph{Tecnalia}, and also a Distinguished Professor at the \emph{University of the Basque Country (UPV/EHU)}\footnote{\url{https://www.ehu.eus/en/en-home}}. His research interests are focused on applied AI (with a focus on \textit{Trustworthy} and \textit{Responsible AI}, \textit{Open-World learning} and \textit{Explainable AI}) for paradigms arising in industry, health, transport, and mobility, among many other fields. He has published more than $470$ journal articles and conference contributions, supervised $19$ PhD thesis, edited $4$ books, and invented $9$ patents. He is a \textit{Senior Member} of IEEE and has been awarded several recognitions for his career. He has been included in the list of the \textit{Top 2\%} most influential AI researchers worldwide by Stanford University (since 2021), and was part of the team that developed the \emph{R\&D\&i} AI strategy for the Spanish government (2019). 
\end{minipage}
\end{spacing}

\bibliographystyle{apalike}
\bibliography{bibliography.bib}

\end{document}